\documentclass{article} 
\usepackage{colm2024_conference}

\usepackage{microtype}
\usepackage{hyperref}
\usepackage{url}
\usepackage{booktabs}
\definecolor{darkblue}{rgb}{0, 0, 0.5}
\hypersetup{colorlinks=true, citecolor=darkblue, linkcolor=darkblue, urlcolor=darkblue}
\usepackage{xcolor}

\usepackage[pdftex]{graphicx}
\usepackage[svgpath=./figures/]{svg}

\usepackage{colortbl}

\usepackage{algorithm}
\usepackage{algpseudocode}

\usepackage{tabularx}
\usepackage{booktabs}
\usepackage[column=O]{cellspace}
\usepackage{multirow}
\usepackage{subcaption}

\usepackage{enumitem}
\setlist[itemize]{leftmargin=*}

\usepackage{amsmath,amsfonts,bm}

\def\eqref#1{equation~\ref{#1}}

\def\1{\bm{1}}

\DeclareMathAlphabet{\mathsfit}{\encodingdefault}{\sfdefault}{m}{sl}
\SetMathAlphabet{\mathsfit}{bold}{\encodingdefault}{\sfdefault}{bx}{n}

\newcommand{\RR}{\mathbb{R}} 
\newcommand{\dataset}[1]{\textsc{#1}}
\newcommand{\camera}[1]{\color{black}{#1}\color{black}} 

\newif\ifcomments
\commentsfalse

\ifcomments
    \usepackage[normalem]{ulem}
    \definecolor{ABpurple}{rgb}{0.8,0.0,0.8}
    \newcommand\abc[1]{\textcolor{ABpurple}{\textsf{\scriptsize[\textbf{AB\@:} #1]}}} 
    \newcommand\abi[1]{\textcolor{ABpurple}{#1}} 
    \newcommand\abm[1]{{\marginparwidth=3cm \marginpar{\raggedright\tiny\textcolor{ABpurple}{\textsf{{\bfseries AB\@:} #1}}}}} 
    \newcommand\abs{\bgroup\markoverwith{\textcolor{ABpurple}{\rule[.4ex]{2pt}{0.8pt}}}\ULon} 
\else
    \newcommand\abc[1]{}
    \newcommand\abi[1]{\ignorespaces}
    \newcommand\abm[1]{}
    \newcommand\abs[1]{#1}
\fi

\title{Instruction-tuning Aligns LLMs to the Human Brain}

\author{%
  Khai Loong Aw,
  Syrielle Montariol\thanks{Equal contribution} ,
  Badr AlKhamissi\footnotemark[1] , 
  Martin Schrimpf\thanks{Equal supervision / senior authors} ,
  Antoine Bosselut\footnotemark[2] \\
  EPFL \\
}

\colmfinalcopy 
\begin{document}

\maketitle

\begin{abstract}
\label{section_abstract}



Instruction-tuning is a widely adopted finetuning method that enables large language models (LLMs) 
to generate output that more closely resembles human responses. 
However, no studies have shown that instruction-tuning actually teaches LLMs to process language in a similar manner as humans.
We investigate the effect of instruction-tuning on aligning LLM and human language processing mechanisms in two ways: (1) \textit{brain alignment}, the similarity of LLM internal representations to neural activity in the human language system, and (2) \textit{behavioral alignment}, the similarity of LLM and human behavior on a reading task.
We assess 25 vanilla and instruction-tuned LLMs on three datasets involving humans reading naturalistic stories and sentences, and find that instruction-tuning generally enhances brain alignment ($\sim$6\%), but has no similar effect on behavioral alignment.
To identify factors underlying this improvement in brain alignment, we compute correlations between brain alignment and various LLM properties, such as model size, problem-solving, and world knowledge understanding.
Notably, we find a strong positive correlation between brain alignment and model size (r = 0.95), as well as performance on tasks requiring world knowledge (r = 0.81). 
Our results demonstrate that instruction-tuning LLMs improves both world knowledge representations and brain alignment, suggesting that the mechanisms that encode world knowledge in LLMs also improve representational alignment to the human brain.

\end{abstract}

\section{Introduction}
\vspace{-5pt}
\label{section_introduction}

Instruction-tuning is a widely adopted method that finetunes large language models (LLMs) on datasets containing many examples of different tasks and their descriptive instructions, enhancing their ability to generalize to previously unseen tasks by learning to follow provided instructions \citep{wang-etal-2022-super}. Despite costing only a small fraction of compute relative to pretraining \citep{chung_scaling_2022}, instruction-tuning yields large performance improvements on reasoning and problem-solving benchmarks. 
This generalization has allowed LLMs to tackle open-world reasoning tasks previously achievable only by humans \citep{zhang_instruction_2023} while only using a few (or zero) task-specific training examples.


In addition to teaching LLMs to understand and follow human instructions, instruction-tuning also improves the ability of LLMs to mimic human-written ground-truth outputs. This fluency allows them to produce more controllable and predictable output that is deemed (1) more desirable to human evaluators \citep{zhang_instruction_2023, chung_scaling_2022, wang_super-naturalinstructions_2022}, (2) more aligned to human values \citep{chia_instructeval_2023}, and (3) more stylistically similar to human outputs \citep{dasgupta_language_2022, safdari_personality_2023}.

Consequently, instruction-tuning yields LLMs more similar to humans in both capability and output resemblance. From a neuroscience perspective, this begs the question: \textbf{Does instruction-tuning make LLMs more functionally similar to the human language system?} Previous work has shown that models with higher task performance are more aligned to the human language system \citep{schrimpf_neural_2021, goldstein_shared_2022, caucheteux_brains_2022}, hitting the estimated noise ceiling\footnote{In fMRI recordings, a noise ceiling for representational similarity can be computed by sampling from the same participant twice, obtaining an upper limit for how well an ideal model could perform, defined by the internal consistency of neural responses and noise level of the data gathering process.} on some datasets. However, there has been no similar study on how instruction-tuning 
affects alignment with the human language system.

In this work, we measure the effect of instruction-tuning on the alignment between language mechanisms in LLMs and the human brain in two ways: (1) \textit{brain alignment}, an ``internal'' metric that assesses how well internal feature representations of LLMs match neural representations in the human language system, and (2) \textit{behavioral alignment}, an ``external'' metric which evaluates the similarity between LLM and human behavioral measurements. In our study, both LLMs and humans are presented with the same language stimuli comprised of naturalistic stories and sentences. For LLMs, we record their internal representations and per-word perplexity. For humans, we use previously recorded brain activity data from functional magnetic resonance imaging (fMRI) experiments and per-word reading times.

To measure brain alignment, we use the Brain-Score \citep{schrimpf_brain-score_2018} linear predictivity metric, assessing how well LLM representations predict human brain activity in response to the same language stimuli \citep{jain_incorporating_2018, toneva_interpreting_2019, schrimpf_neural_2021, oota_deep_2023}, using data from three neural datasets:  \citet{pereira_toward_2018}, \citet{blank_functional_2014}, and \citet{wehbe_simultaneously_2014}. To evaluate behavioral alignment, we use a benchmark in Brain-Score which calculates the Pearson correlation between LLM per-word perplexity and human per-word reading times from the \citet{futrell_natural_2018} dataset. Perplexity for LLMs and reading times for humans offer insights into comprehension difficulty \citep{ehrlich_contextual_1981, hale_probabilistic_2001, smith_effect_2013}, allowing us to examine whether LLMs match humans in their patterns of words and sentences they find challenging or surprising. Because models vary in their brain and behavioral alignment across different architectures and training objectives \citep{schrimpf_neural_2021}, we estimate the effect of instruction-tuning by evaluating 8 vanilla LLMs and 17 LLMs that were further instruction-tuned, and report a significant increase in brain alignment due to instruction-tuning.


To investigate \textit{why} instruction-tuning increases alignment to human brain activity, we study the relationships between brain alignment and various LLM properties. Specifically, we compute Pearson correlations between an LLM's brain alignment and its properties, including next-word prediction (NWP) ability, model size, a range of problem-solving abilities, and world knowledge spanning different domains. We evaluated the latter two properties using the Big-Bench Hard benchmark (BBH; \citealp{suzgun_challenging_2022}) and the Massive Multi-task Language Understanding benchmark (MMLU; \citealp{hendrycks_measuring_2021}), respectively. 

\begin{figure}[!ht]
    \includegraphics[width=1.0\linewidth]{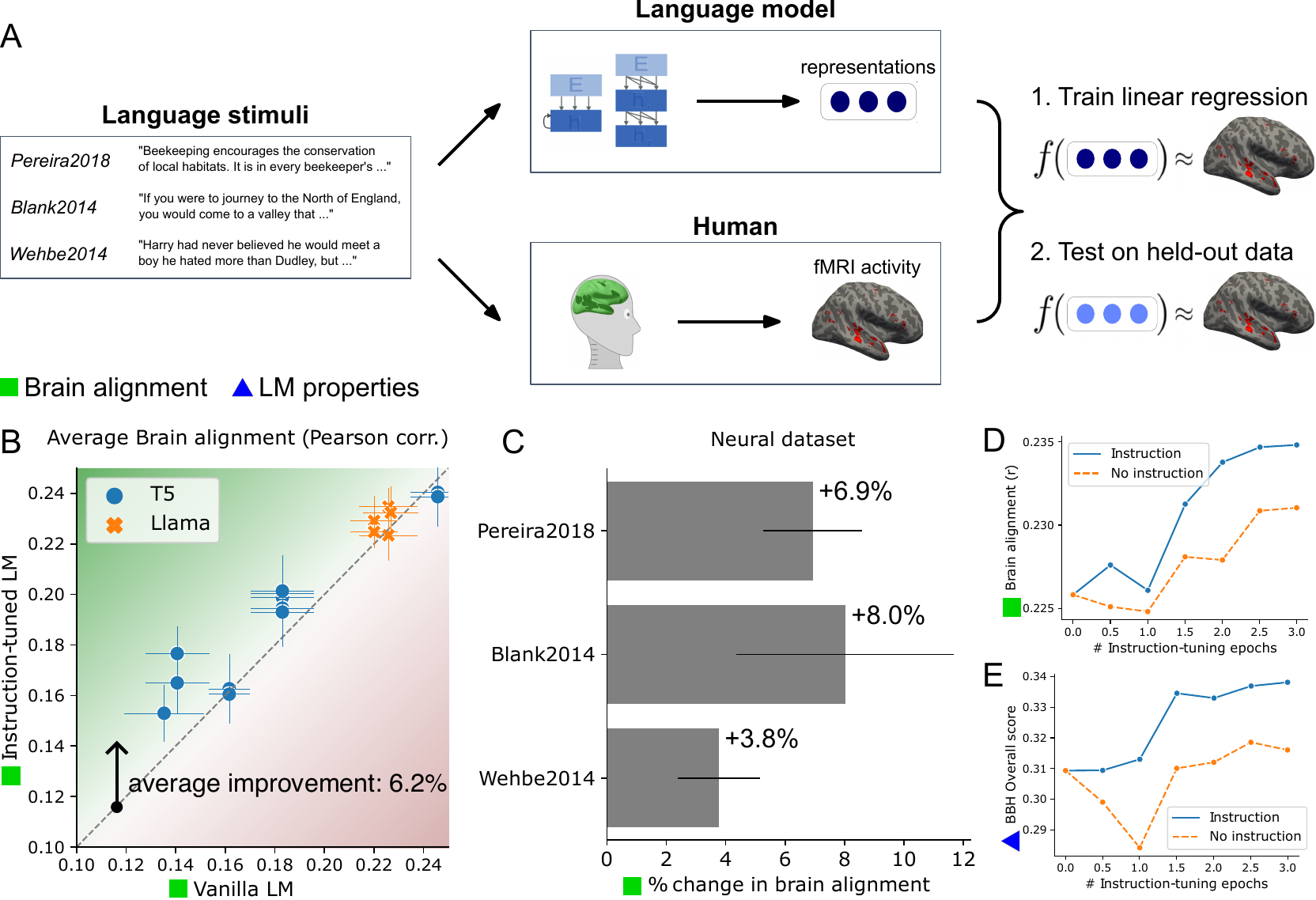}
    \caption{
        \textbf{Instruction-tuning aligns LLM representations to human brain activity.}
        \textbf{(A)}
        The same language stimuli are presented to LLMs and human participants. Next, we fit a linear regression model from LLM layer activations to fMRI responses in the human language system. We apply this linear model to predict held-out fMRI responses from the original corpus of recordings, and compute brain alignment as the Pearson correlation between the predicted and actual fMRI responses. 
        We evaluate 25 vanilla and instruction-tuned LLMs with sizes between 77M and 33B parameters.
        We compute the average across 3 neural datasets of humans reading naturalistic stories and sentences: \dataset{Pereira2018}, \dataset{Blank2014}, and \dataset{Wehbe2014}.
        \textbf{(B)}
        Instruction-tuning improves average brain alignment by 6.2\% on average.
        Each point above the identity line is an instruction-tuned LLM with greater brain alignment than its vanilla version. Error bars, here and elsewhere, represent median absolute deviation over human participants.
        \textbf{(C)}
        Instruction-tuning improves average brain alignment on all three datasets.
        \textbf{(D)}
        We instruction-tune LLaMA-7B on the Alpaca dataset (``Instruction'' model) and train an ablation model with the same data, but without the instruction in each training sample (``No Instruction" model).
        Our results show that brain alignment improvements are due to both (1) training data (present in both models) and (2) training LLMs to understand and follow instructions (present only in the first model).
    }
    \label{fig:result_instruction_tuning_BA}
\end{figure}

We report three major findings:
\begin{enumerate}
    [nolistsep]
    \itemsep 0.5em 

    \item Instruction-tuning generally aligns LLM representations to human brain activity, increasing brain alignment by $6.2$\% on average (Figure \ref{fig:result_instruction_tuning_BA}).
  
    \item Investigating the factors underlying LLM-brain alignment, we find that brain alignment is strongly correlated with world knowledge (r = $0.81$) and model size (r = $0.95$) (Figure \ref{fig:result_BA_correlates_WK_MS}). 

    \item Surprisingly, our results indicate that instruction-tuning LLMs generally does not enhance behavioral alignment with human reading times. Furthermore, behavioral alignment on this dataset is poorly correlated with all other measures we investigate, including task performance and model size (Figure \ref{fig:result_BehavA}).
\end{enumerate}

\section{Background \& Related Work}
\vspace{-5pt}
\label{section_related_work}

\textbf{Effect of Instruction-tuning on LLMs.} \:
Instruction-tuning is an effective method for enhancing LLM capability and controllability. It trains LLMs using pairs of human instructions and desired outputs. The benefits of instruction-tuning can be categorized into three key aspects \citep{zhang_instruction_2023}: (1) it bridges the disparity between the pretraining objective of LLMs (next-word prediction) and the goal of accurately following human instructions, (2) it achieves greater control and predictability of model behavior compared to standard LLMs, allowing researchers to make them more similar to humans in both capability and output similarity \citep{chia_instructeval_2023, dasgupta_language_2022, safdari_personality_2023}, and (3) it often costs only a small fraction of compute relative to pretraining, enabling LLMs to swiftly adapt to target domains \citep{chung_scaling_2022}. We contribute to this research area from a neuroscience perspective, by studying whether instruction-tuning makes LLMs more aligned to the human language system in terms of brain and behavioral alignment.


\textbf{Effect of Finetuning on Brain alignment.} \:
Prior works have studied how finetuning affects LMs' alignment to human brain activity.
These include finetuning on a wide range of downstream NLP tasks \citep{oota_neural_2022}, finetuning to summarize narratives \citep{aw_training_2023}, and finetuning to directly predict brain activity recordings \citep{schwartz_inducing_2019}. These studies aim to use brain alignment to study how finetuning affects LMs and their representations.
Our work builds on this line of research by demonstrating that instruction-tuning aligns LLM representations to human brain activity. We also investigate why instruction-tuned LLMs align to brain activity by testing the correlation of brain alignment with various world knowledge domains and problem-solving abilities.

\textbf{LM properties linked to Brain alignment.} \:
One exciting research area focuses on disentangling the contribution of various LM properties towards brain alignment. 
These include studying how brain alignment is driven by next-word prediction ability \citep{schrimpf_neural_2021, caucheteux_brains_2022}, multi-word semantics \citep{merlin_language_2022}, performance on various NLP tasks \citep{oota_neural_2022}, and model size \citep{antonello_scaling_2023}.
To contribute to this growing body of work, we use instruction-tuned LLMs. They are especially useful as they have been trained to respond to a standard question-answer format, allowing us to evaluate LLMs on a wide array of tasks and in a more fine-grained manner.
We expand this area of research by demonstrating that world knowledge is a key property underlying LLM-brain alignment.

\section{Language Models}
\vspace{-5pt}
\label{method_language_models}
We evaluate the brain alignment of 25 large language models (LLMs) from two model families: T5 \citep{raffel_exploring_2020} and LLaMa \citep{touvron_llama_2023}. 
T5 models are encoder-decoder LLMs pre-trained using a masked infilling objective on the Colossal Common Crawl Corpus (C4), a corpus of 356 billion tokens, and then further finetuned on a multi-task mixture of unsupervised and supervised tasks converted into a text-to-text format. We use all five T5 models, with sizes between 77M to 11B. 
LLaMA models \citep{touvron_llama_2023} are decoder-only LLMs trained on 1.6 trillion tokens from a mixture of corpora including C4, English CommonCrawl, Wikipedia, GitHub. We use the 7B, 13B, and 33B versions.

For the instruction-tuned variants of T5 models, we use a variety of models finetuned on FLAN ($15$M examples for 1,836 different tasks accompanied by instructions, \citealp{chung_scaling_2022}), Alpaca ($52$K instruction-following examples generated with methods inspired by Self-Instruct, \cite{selfinstruct}, \citealp{alpaca}), and GPT4ALL ($437$K instruction-following examples generated with GPT-3.5-turbo, \citealp{gpt4all}) datasets. For the instruction-tuned variants of LLaMa, we use Vicuna's $7$B, $13$B, and $33$B models \citep{vicuna2023}, which are finetuned on user-shared conversations. We also use  StableVicuna-13B, which further refines Vicuna-13B using reinforcement learning from human feedback (RLHF) \citep{Ouyang2022TrainingLM} on a range of conversational and instructional datasets. We also use the $7$B version of Alpaca \citep{alpaca}. We provide more details in Appendix \ref{appendix_LM_counts}. 

We also tested 8 models from the GPT2 family; four vanilla models (Small, Medium, Large, XL) and their corresponding versions further instruction-tuned on Alpaca. However, the models performed close to random on reasoning and world knowledge benchmarks (BBH, MMLU). We hypothesize this is due to their size (all $<$ 1.5B parameters), making fine-tuning on a relatively small instruction-following dataset insufficient. Moreover, GPT2 was not originally pretrained to perform instruction-following tasks (unlike T5 models). We include their MMLU, BBH, brain and behavioral alignment results in Appendix \ref{appendix_results_BA}, \ref{appendix_results_mmlu_bbh}, and \ref{appendix_results_BehavA}.

\begin{table}[t]
    \centering

    {
    \begin{tabular}{lll}
    \toprule
    Instruction & Input & Output \\
    \midrule
    \multirow{2}{*}
    \textnormal{``Write a short paragraph} & \textnormal{``The importance of} & \textnormal{``The use of renewable energy} \\
    about the given topic." & \textnormal{using renewable energy."} & {is growing..." (paragraph)} \\
    \bottomrule
    \end{tabular}

    \caption{
        \textbf{Example of Instruction-tuning training data format: (Instruction, Input, Output)} from the Alpaca dataset \citep{alpaca}. The input field is optional for certain instructions.
    }
    \label{table:instruction_tuning_data_format}
    }
\end{table}

\section{Brain Alignment}
\vspace{-5pt}
\label{section_brain_alignment}

Brain alignment refers to the method of evaluating the similarity between LLM representations and human brain activity (Figure \ref{fig:result_instruction_tuning_BA}A and Appendix \ref{appendix_methods_brain_alignment}). This relies on fMRI recordings of human subjects while they read language stimuli on potentially any topic \citep[here:][]{pereira_toward_2018,blank_functional_2014,wehbe_simultaneously_2014}. The same language stimuli from prior brain recordings are provided as input to LLMs, whose intermediate layer activations are extracted as their representations of the language stimuli. We follow a general approach previously used in several works \citep{schrimpf_brain-score_2018,jain_incorporating_2018, toneva_interpreting_2019, schrimpf_neural_2021, oota_deep_2023, aw_training_2023}. Specifically, we use the linear predictivity metric implemented in Brain-Score \citep{schrimpf_brain-score_2018}, which fits a linear regression model from LLM layer activations to fMRI responses in the human language system. We then apply this linear model to predict held-out fMRI responses from the original corpus of recordings, and compute brain alignment as the Pearson correlation between the predicted and actual fMRI responses. For each LLM, we evaluate its brain alignment for every layer (e.g., LLaMA-7B has 32 layers), and use the highest value as the LLM's brain alignment value, following \cite{schrimpf_brain-score_2018}.

\textbf{Datasets} \: We use three fMRI datasets to measure the brain alignment of LLMs. Each dataset involves a different set of human participants and a different set of language stimuli.

\dataset{Pereira2018} (experiments 2 and 3 from \citealp{pereira_toward_2018}): In experiment 2, 9 participants read 384 sentences taken from 96 text passages. In experiment 3, 6 participants read 243 sentences from 72 text passages. Each sentence was displayed for 4 seconds on a screen.

\dataset{Blank2014} \citep{blank_functional_2014}: The data consists of fMRI recordings of 5 human participants listening to 8 naturalistic stories from the Natural Stories Corpus \citep{futrell_natural_2018}.

\dataset{Wehbe2014} \citep{wehbe_simultaneously_2014}: The data includes fMRI recordings of 8 human participants reading chapter 9 of the book \textit{Harry Potter and the Sorcerer's Stone} \citep{rowling1998harry}. Participants read the chapter at a fixed interval of one word every 0.5 seconds.

\subsection{Instruction-tuning aligns LLM representations to human brain activity}
\label{subsection_result_IT_BA}
\vspace{-5pt}
First, we study the effect of instruction-tuning on LLM brain alignment. We compute each LLM's average brain alignment as the mean of its brain alignment on the 3 neural datasets. We find that instruction-tuning improves brain alignment by an average of 6.2\% across tested LLMs (Figure \ref{fig:result_instruction_tuning_BA}B).
This holds across all three neural datasets, with improvements of +6.9\% on \dataset{Pereira2018}, +8.0\% improvement on \dataset{Blank2014}, and +3.8\% on \dataset{Wehbe2014} (Figure \ref{fig:result_instruction_tuning_BA}C). Moreover, a smaller instruction-tuned model can attain higher brain alignment than a larger vanilla model from the same family (e.g., Alpaca-7B v.s. LLaMa-13B, see Appendix \ref{appendix_results_BA}). However, some of the larger models (T5-XL, LLaMA) have minimal or no increases, possibly because the vanilla models are already close to the noise ceiling (Appendix \ref{appendix_results_BA}).

Next, to longitudinally study how instruction-tuning aligns LLM representations to brain activity, we separately instruction-tune a LLaMA-7B model on the Stanford Alpaca instruction dataset \citep{alpaca} for 3 epochs. By evaluating checkpoints regularly during training, we find that instruction-tuning progressively improves brain alignment (Figure \ref{fig:result_instruction_tuning_BA}D, Appendix \ref{appendix_results_IT_llama_7b}). We also disambiguate the effect on brain alignment of (1) the instruction-following ability gained from instruction-tuning and (2) added training data. We fine-tune LLaMA-7B with the same process and data, but remove the instruction from each training sample. Brain alignment of this ablated model increases during fine-tuning but stays lower than its instruction-following counterpart (Figure \ref{fig:result_instruction_tuning_BA}D), indicating that brain alignment improvements are due to both factors. 
\camera{
%
%
In Appendix \ref{appendix_ba_results_rsa_cka}, we also report preliminary results using recent models from the Gemma \citep{gemmateam2024gemmaopenmodelsbased} and LLaMA-2 \citep{touvron2023llama2openfoundation} families  and observe similar trends. Further, these trends hold across different similarity metrics besides Linear predictivity, e.g., RSA and CKA (Appendix \ref{appendix_ba_results_rsa_cka}). 
}

\begin{figure}[t]
    \includegraphics[width=1.0\linewidth]{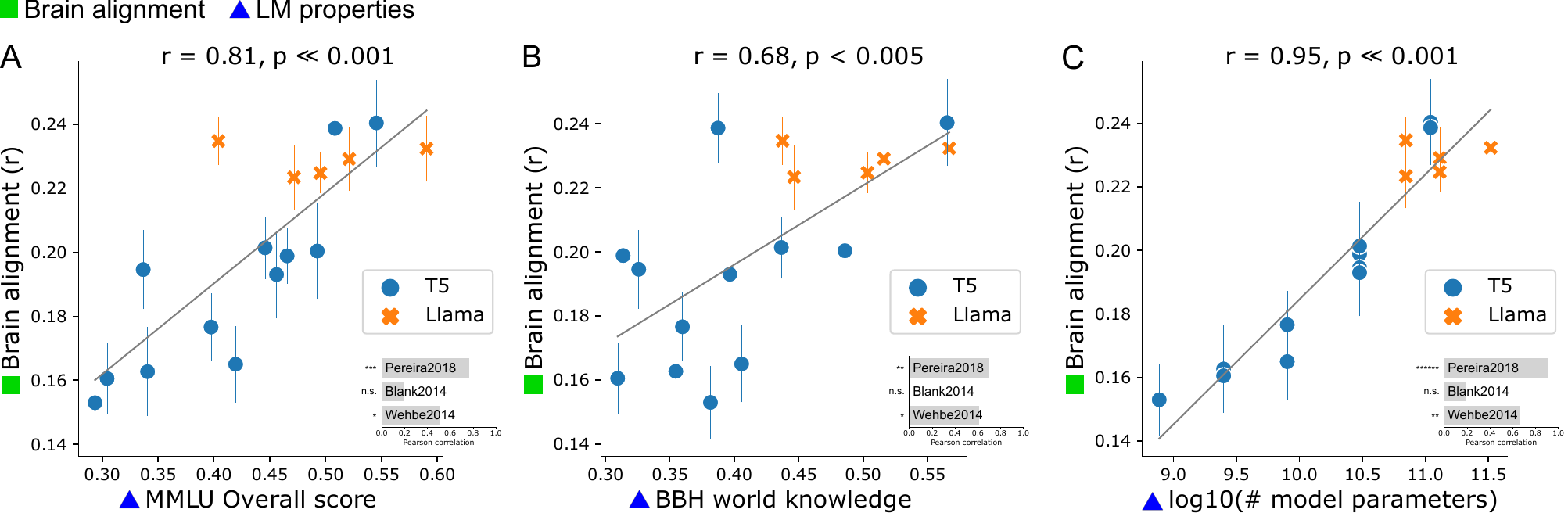}
    \caption{
        \textbf{World knowledge and model size are important factors underlying LLM-brain alignment.}
        To identify factors underlying brain alignment, we test Pearson correlations between brain alignment and various LLM properties, such as model size, world knowledge in various domains (MMLU benchmark), and various types of problem-solving abilities (BBH benchmark). 
        In the figure, insets display results on individual datasets, with stars reflecting statistical significance (n.s. = p $>$ 0.05, * = p $<$ 0.05, ** = p $<$ 0.005, etc.)
        \textbf{(A)} 
        Brain alignment is significantly and strongly correlated with world knowledge as evaluated by the MMLU Overall score (r = 0.81), which reports the mean performance across all MMLU subjects.
        \textbf{(B)} 
        Brain alignment is significantly and strongly correlated with performance on the world knowledge task category in BBH (r = 0.68).
        \textbf{(C)} 
        Brain alignment is significantly and strongly correlated with model size (logarithm of number of parameters) (r = 0.95). In Appendix \ref{appendix_results_corrs_BA}, we provide a larger version of this figure with labels for each data point.
    }
    \label{fig:result_BA_correlates_WK_MS}
\end{figure}

\begin{table}[t]
    \centering

    {
    \renewcommand{\arraystretch}{0.8}
    \begin{tabular}{l r r r r}
    \toprule
    \multirow{2}{*}{\textbf{Benchmark, Task category}}& \textbf{Correlation (\textit{r}) to} & \textbf{Adjusted} & \textbf{No.} & \textbf{Avg. Model} \\
    & \textbf{Brain Alignment} &  \textit{p}-\textbf{value} & \textbf{tasks} & \textbf{Accuracy}\\
    \midrule
    MMLU, Overall Score    & \textbf{0.809} & \textbf{$<$ 0.0005} & 57 & 0.36 \\
    MMLU, STEM             & \textbf{0.792} & \textbf{$<$ 0.0005} &  18 & 0.28 \\
    MMLU, Humanities       & \textbf{0.791} & \textbf{$<$ 0.0005} &  13 & 0.34 \\
    MMLU, Social Sciences  & \textbf{0.807} & \textbf{$<$ 0.0005} &  12 & 0.41 \\
    MMLU, Others           & \textbf{0.809} & \textbf{$<$ 0.0005} &  14 & 0.40 \\
    \midrule
    BBH, Overall score             & 0.384 & 0.18 &  23 & 0.28 \\
    BBH, Algorithmic reasoning     & 0.194 & 0.56 &  8 & 0.22 \\
    BBH, Language understanding    & 0.163 & 0.59 &  3 & 0.43 \\
    BBH, World knowledge           & \textbf{0.679} & \textbf{$<$ 0.005} &  5 & 0.36 \\
    BBH, Multilingual reasoning    & -0.035 & 0.90 &  1 & 0.19 \\
    BBH, Others                    & 0.478 & 0.08 &  6 & 0.27 \\
    \bottomrule
    \end{tabular}
    

    \caption{
        \textbf{Brain alignment strongly correlates with world knowledge in all subject domains in MMLU and the world knowledge category in BBH.} Brain alignment is not significantly correlated with other problem-solving abilities in BBH (e.g., algorithmic or multilingual reasoning).
        We obtain p-values after performing false discovery rate (FDR) correction. 
    }
    \label{table:result_BA_correlates_MMLU_BBH}
    }
\end{table}

\subsection{Factors underlying LLM-brain alignment}
\label{subsection_result_WK_MS}
\vspace{-5pt}

To study why LLMs and human brains align in their representations, we compute the Pearson correlation between LLM brain alignment and various properties of LLMs:  performance on a benchmark testing various reasoning abilities (BBH; \citealp{suzgun_challenging_2022}), performance on a benchmark testing world knowledge in various domains (MMLU; \citealp{hendrycks_measuring_2021}), language modeling ability, and model size.

\textbf{MMLU and BBH.} \: MMLU is designed to measure the world knowledge of LLMs across many subject domains. It contains 57 tasks, categorized into four world knowledge subject domains: STEM, Humanities, Social Sciences, and Others (a broad category ranging from finance to marketing to professional medicine). BBH contains 23 tasks, grouped into four types of problem-solving: Algorithmic and Multi-Step Arithmetic Reasoning; Natural Language Understanding; Use of World Knowledge; and Multilingual Knowledge and Reasoning. For both benchmarks, we follow the category classifications from the original papers. We perform the evaluations using the \texttt{instruct-eval} repository\footnote{\url{https://github.com/declare-lab/instruct-eval}} with default settings (3-shots for BBH, 5-shots for MMLU) and preset prompts.
We measure the Pearson correlation (and its p-value) between LLM-brain alignment and performance in each category of MMLU and BBH. We obtain p-values after false discovery rate (FDR) correction. 

\textbf{World Knowledge.} We find that brain alignment is significantly and strongly correlated with world knowledge. Brain alignment is highly correlated with the MMLU Overall score (r = 0.81, p $<$ 0.001, Figure \ref{fig:result_BA_correlates_WK_MS}A), which reports the mean performance across all world knowledge subject domains on MMLU. Similarly, brain alignment is also strongly correlated with performance on tasks in the world knowledge category of BBH (r = 0.68, p $<$ 0.005; Figure \ref{fig:result_BA_correlates_WK_MS}B). Notably, brain alignment is poorly correlated with all other dimensions of BBH (
see Table \ref{table:result_BA_correlates_MMLU_BBH}), though this could also be due to limitations of the tested models, 
as indicated by their low raw performance scores on some tasks.  
Overall, our results provide a strong signal that more accessible representations of world knowledge are a key factor in aligning LLM representations to human brain activity.

\textbf{Language Modeling Ability.} \: Prior works have shown correlations between brain alignment and next-word prediction (NWP) ability \citep{caucheteux_brains_2022, schrimpf_neural_2021}. We find similar results for correlation between brain alignment and NWP loss (r = -0.54, p $<$ 0.05, Appendix \ref{appendix_results_corrs_BA}). Interestingly, the strength of the correlation is weaker than that between brain alignment and world knowledge performance (r = 0.81). This suggests that world knowledge understanding is a better predictor of brain alignment than NWP ability. 

\textbf{Model Size.} \: Finally, we find that brain alignment is significantly and strongly correlated with model size (r = 0.95, p $<$ 0.001, Figure \ref{fig:result_BA_correlates_WK_MS}C), as measured by the logarithm of the number of model parameters. \citet{schrimpf_neural_2021} observe such a pattern for language models, and we find the pattern holds for instruction-tuned models, and models trained at a larger scale than their study (7B+ parameters).
However, model size alone does not determine brain alignment. Our results show that smaller instruction-tuned LLMs can have greater brain alignment than larger vanilla models. For example, LLaMA-13B obtains brain alignment of 0.220, Vicuna-13B obtains 0.229, LLaMA-33B obtains 0.227, and Vicuna-33B obtains 0.232. Hence, Vicuna-13B has greater brain alignment than LLaMA-33B, due to instruction-tuning, despite being less than 40\% its size. We observe a similar trend in another four models: T5-base, Flan-T5-base, T5-large, Flan-T5-large.
Also, prior works showed that large random models achieve poor brain alignment \citep{schrimpf_neural_2021}. These results demonstrate there are LLM properties aside from model size that contribute significantly to brain alignment. 

\textbf{Neural datasets.} \: Brain alignment is strongly correlated with world knowledge and model size on \dataset{Pereira2018} and \dataset{Wehbe2014}, but less strongly correlated on \dataset{Blank2014}, possibly because \dataset{Blank2014} has fewer participants (N = 5) leading to greater noise in the results.

\section{Behavioral Alignment}
\label{section_behavioral_alignment}
\vspace{-5pt}

\begin{figure}[t]
    \includegraphics[width=1.0\linewidth]{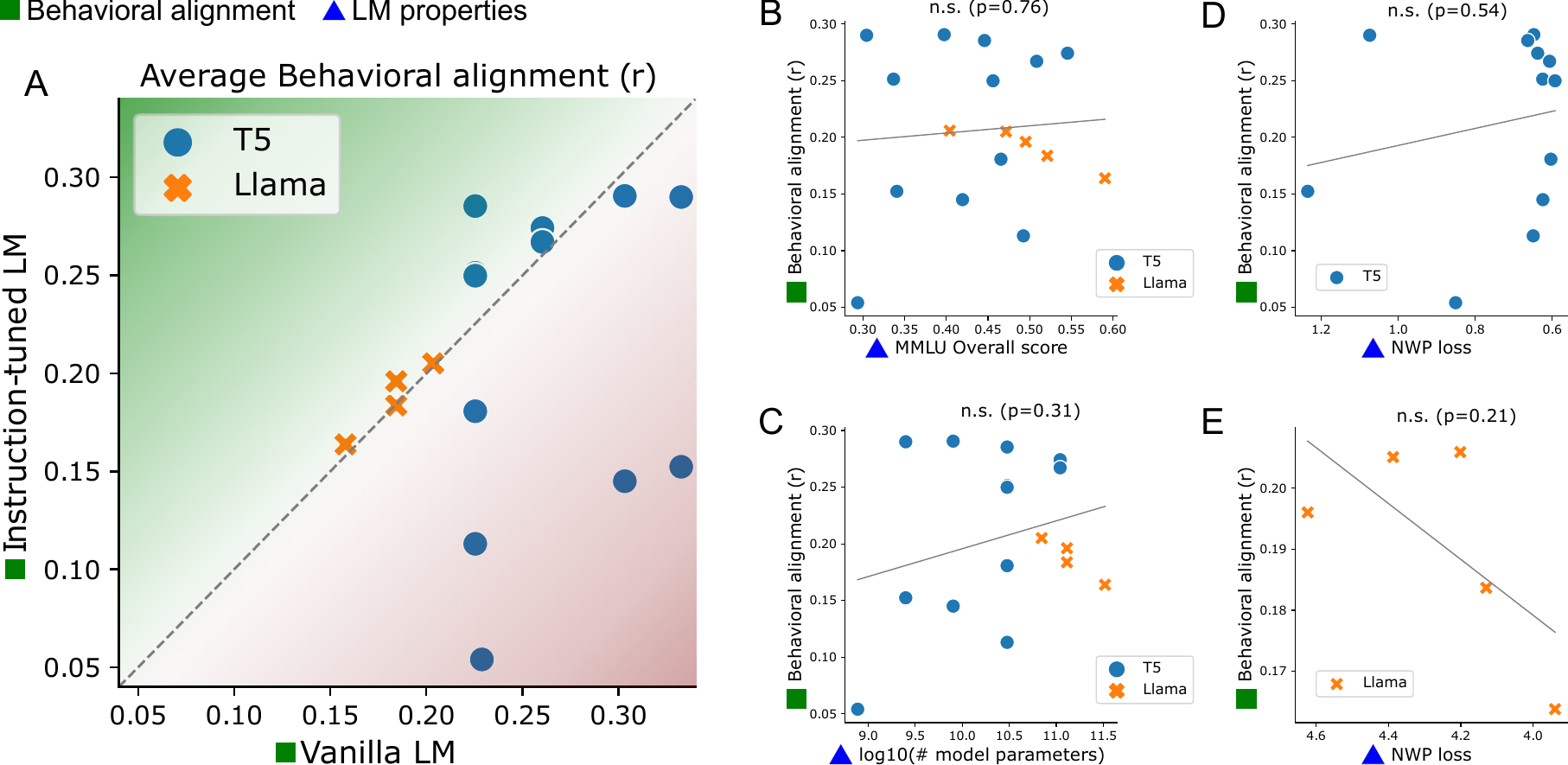}
    \caption{
        \textbf{Instruction-tuning LLMs generally does not improve behavioral alignment to human reading times. Furthermore, behavioral alignment correlates poorly with all other tested measures:}
        world knowledge, model size, and next-word prediction (NWP) ability.   
        To compute behavioral alignment, we use the \dataset{Futrell2018} benchmark in Brain-Score. The same language stimuli (naturalistic stories) are presented to LLMs and human participants. We then compute the Pearson correlation between per-word LLM perplexity and human reading times as the behavioral alignment.
        \textbf{(A)} 
        Instruction-tuning does not generally improve behavioral alignment. 
        Furthermore, behavioral alignment is poorly and not significantly correlated with all other measures: 
        \textbf{(B)} world knowledge (p = 0.76), 
        \textbf{(C)} model size (p = 0.31), 
        \textbf{(D)} NWP loss for T5 models (p = 0.54), and 
        \textbf{(E)} NWP loss for LLaMA models (p = 0.21). 
        In Appendix \ref{appendix_results_corrs_BehavA}, we provide a larger, labeled version of this figure.
    }
    \label{fig:result_BehavA}
\end{figure}

In the previous section, we show that instruction-tuning aligns the internal representations of LLMs to human \textit{brain activity} (Section \ref{subsection_result_IT_BA}). In this section, we explore whether instruction-tuning also aligns LLM behavior to human \textit{behavior}.

Following the approach of \citet{schrimpf_neural_2021} implemented in the Brain-Score package \citep{schrimpf_integrative_2020}, we measure behavioral alignment by evaluating the similarity between LLM per-word perplexity and human per-word reading times given the same language stimuli (Appendix \ref{appendix_methods_behavioral_alignment}). We use the self-paced reading times dataset from \citet{futrell_natural_2018}, consisting of the reading times of 179 human participants recorded while they were visually presented with 10 naturalistic stories. We provide language stimuli from this data as input to LLMs and 
evaluate behavioral alignment by computing the Pearson correlation between per-word LLM perplexity and per-word human reading times.

\camera{
One potential concern in the behavioral alignment methodology is that human reading time is influenced by word length; longer words may take more time to read, even if they are not more challenging or surprising. Following \cite{schrimpf_neural_2021}, we address this by computing the surprisal of a multi-token word as the sum of the surprisals of its individual tokens. Importantly, we use consistent tokenizers when comparing vanilla LLMs against their instruction-tuned versions, thereby controlling for the effect of word length when evaluating the impact of instruction-tuning on behavioral alignment.
}


\subsection{{Instruction-tuning generally does not improve behavioral alignment}}
\vspace{-5pt}

Using the same models as in Section \ref{section_brain_alignment}, we compare the behavioral alignment of each instruction-tuned LLM against its vanilla counterpart. Our results indicate instruction-tuning generally does not improve behavioral alignment to human reading times (Figure~\ref{fig:result_BehavA}A). For half of the LLMs tested, it results in no change or reduced behavioral alignment. 
\camera{
Our results align with recent studies \citep{gao-etal-2023-roles, steuer-etal-2023-large} that show improving LLM performance does not necessarily improve alignment with human reading behavior. 
}


\subsection{Factors underlying behavioral alignment}
\vspace{-5pt}
Next, we test the correlation between LLM behavioral alignment and model size, next-word prediction ability, various reasoning abilities (measured by BBH), and world knowledge across various domains (measured by MMLU). Contrary to our findings on the brain alignment correlations with model size and world knowledge (Section \ref{subsection_result_WK_MS}), we find that LLM behavioral alignment is not correlated with these factors: world knowledge (p = 0.76, Figure \ref{fig:result_BehavA}B), model size (p = 0.31, Figure \ref{fig:result_BehavA}C), next-word prediction loss for T5 models (p = 0.54, Figure \ref{fig:result_BehavA}D), and next-word prediction loss for LLaMA models (p = 0.21, Figure \ref{fig:result_BehavA}E). 


\section{Discussion}
\label{section_discussion}
\vspace{-5pt}



\subsection{{Implications for NLP: Building LLMs}}
\label{subsection_discussion_NLP}
\vspace{-5pt}

\textbf{Using brain alignment for LLM interpretability.} 
Prior works have used human brain activity to interpret models \citep{dong_interpreting_2023} and build more performant models \citep{dapello_simulating_2020,safarani_towards_2021,dapello_aligning_2022}. Instruction-tuning has emerged as a breakthrough technique to improve LLM abilities and quality of outputs, and allows LLMs to adapt to new tasks with minimal task-specific training. 
However, how instruction-tuning alters LLM internal representations to achieve these improvements remains an open question. 
Brain activity provides a neuroscientific angle to investigate this question. 
Our results show that instruction-tuning improves both performance on world knowledge benchmarks and human brain alignment, 
suggesting that the mechanisms that encode world knowledge in LLMs also align LLM representations to the human brain. 



\subsection{{Implications for Neuroscience: Studying LLM-Human Alignment}}
\label{subsection_discussion_NeuroNLP}
\vspace{-5pt}

\textbf{Instruction-tuned LLMs are useful for studying LLM properties underlying brain and behavioral alignment.} \:
To identify why LLM and human brains share representational similarities, prior work has mostly focused on high-level properties such as model size \citep{antonello_scaling_2023}, as well as external behaviors such as predicting missing words \citep{schrimpf_neural_2021, caucheteux_brains_2022}.
However, a key to understanding these similarities is to identify internal properties of LLMs that underlie brain alignment, including the amount of knowledge LLMs encode, e.g., factual \citep{alkhamissi_review_2022} and commonsense \citep{Bosselut2019COMETCT,sap_commonsense_2020}.
Our work is the first to show that we can harness instruction-tuned LLMs for this purpose. Because they have been trained to respond to a general instruction format, we can evaluate LLMs on diverse tasks in a fine-grained manner, allowing the study of LLM properties both internal (e.g., knowledge) and external (e.g., behavior), and how they correlate with brain and behavioral alignment.

\textbf{World knowledge shapes language comprehension and brain activity.} \:
Our results show that world knowledge is a key factor in LLM brain alignment. LLMs demonstrating greater world knowledge across all tested subject domains have representations more similar to the human brain.
This suggests that world knowledge influences human brain activity and shapes the language comprehension systems in our brain. 

\subsection{{Limitations and Future Work}}
\label{subsection_limitations}
\vspace{-5pt}

\textbf{Examining more dimensions of behavior.} \:
Our work and many prior works compare LLM and human next-word surprisal on reading tasks \citep{wilcox_predictive_2020, schrimpf_neural_2021, eghbal_a_hosseini_artificial_2023}, which evaluates only a single behavioral dimension for LLMs (per-word perplexity) and humans (reading times). 
For the models we tested, behavioral alignment is not significantly correlated with model size, world knowledge, or next-word prediction ability.
While next-word prediction performance correlates with alignment to human reading times across many LMs \citep{schrimpf_neural_2021}, this trend does not hold up in recent transformer models \citep{oh_why_2023}, having a surprising negative correlation with parameter count \citep{oh_comparison_2022}.
In the future, we hope to evaluate LLMs on a wider range of behavioral dimensions (e.g., \citealp{van-duijn-etal-2023-theory, koo2023benchmarking, kauf2024comparing}) to more holistically evaluate 
LLM-human behavioral alignment.

\textbf{Brain alignment datasets with humans performing diverse tasks.} \:
We study brain alignment using neural datasets of humans reading naturalistic stories and sentences in English.
It would be interesting to study brain alignment to human participants attempting the BBH and MMLU benchmarks, but this data unfortunately does not exist.
This may explain why brain alignment is not significantly correlated with many categories of problem-solving on BBH, e.g., algorithmic reasoning.
In the future, we hope to study brain alignment with human participants performing more diverse sets of tasks, e.g., reading computer program code \citep{ivanova_comprehension_2020}.
This can identify more factors underlying LLM-brain alignment, and provide insights into how brain activity and the human language system may be shaped by various forms of problem-solving.
Furthermore, some of the larger models exceed the noise ceiling estimates in our neural datasets (Appendix \ref{appendix_results_BA}), highlighting the need for more neural datasets and better ways of computing noise ceiling estimates.

\section{Conclusion}
\label{section_conclusions}
\vspace{-5pt}


We investigate whether instruction-tuning improves the alignment of LLMs to the human language system. We evaluate 25 LLMs with parameter sizes ranging from 77 million to 33 billion, across three neural datasets of humans reading naturalistic stories and sentences. We find that instruction-tuning generally improves the alignment of LLM representations to brain activity. 
Exploring the factors underlying LLM-brain alignment, we discover that world knowledge and model size are key determinants of brain alignment. This suggests that world knowledge helps shape representations in the human language system, and highlights the importance of integrating world knowledge in developing future LLMs.

\clearpage



\section{Reproducibility Statement}
\label{section_reproducibility}

We provide our full results in Appendices \ref{appendix_results_BA}, \ref{appendix_results_mmlu_bbh}, and \ref{appendix_results_BehavA}. We consider this important as our experiments may be computationally expensive to replicate. We investigated 25 LLMs, with the largest models having 33B parameters, on many datasets: brain alignment (\dataset{Pereira2018}, \dataset{Blank2014}, \dataset{Wehbe2014}), behavioral alignment (\dataset{Futrell2018}), next-word prediction (WikiText-2) and other evaluations (BBH, MMLU).

All models and code repositories used in our study are open-source and their corresponding links are provided in Appendix \ref{appendix_LM_urls} and \ref{appendix_code_repos} for full transparency and reproducibility. To calculate brain and behavioral alignment, we used the Brain-Score repository (\url{www.github.com/brain-score/language}), a publicly accessible resource for conducting these assessments. We encourage researchers interested in replicating our findings to refer to the provided links and consult the Brain-Score repository for further details on datasets and the evaluation process. To evaluate LLMs on BBH and MMLU, we utilize the widely-used \texttt{instruct-eval} repository (\url{https://github.com/declare-lab/instruct-eval}) with default settings.



\bibliography{ref_manual, ref_zotero}

\begin{thebibliography}{56}
\providecommand{\natexlab}[1]{#1}
\providecommand{\url}[1]{\texttt{#1}}
\expandafter\ifx\csname urlstyle\endcsname\relax
  \providecommand{\doi}[1]{doi: #1}\else
  \providecommand{\doi}{doi: \begingroup \urlstyle{rm}\Url}\fi

\bibitem[AlKhamissi et~al.(2022)AlKhamissi, Li, Celikyilmaz, Diab, and Ghazvininejad]{alkhamissi_review_2022}
Badr AlKhamissi, Millicent Li, Asli Celikyilmaz, Mona Diab, and Marjan Ghazvininejad.
\newblock A {Review} on {Language} {Models} as {Knowledge} {Bases}, April 2022.
\newblock URL \url{http://arxiv.org/abs/2204.06031}.
\newblock arXiv:2204.06031 [cs].

\bibitem[Anand et~al.(2023)Anand, Nussbaum, Duderstadt, Schmidt, and Mulyar]{gpt4all}
Yuvanesh Anand, Zach Nussbaum, Brandon Duderstadt, Benjamin Schmidt, and Andriy Mulyar.
\newblock Gpt4all: Training an assistant-style chatbot with large scale data distillation from gpt-3.5-turbo.
\newblock \url{https://github.com/nomic-ai/gpt4all}, 2023.

\bibitem[Antonello et~al.(2023)Antonello, Vaidya, and Huth]{antonello_scaling_2023}
Richard Antonello, Aditya Vaidya, and Alexander~G. Huth.
\newblock Scaling laws for language encoding models in {fMRI}, May 2023.
\newblock URL \url{http://arxiv.org/abs/2305.11863}.
\newblock arXiv:2305.11863 [cs].

\bibitem[Aw \& Toneva(2023)Aw and Toneva]{aw_training_2023}
Khai~Loong Aw and Mariya Toneva.
\newblock Training language models to summarize narratives improves brain alignment.
\newblock In \emph{The Eleventh International Conference on Learning Representations}, 2023.

\bibitem[Blank et~al.(2014)Blank, Kanwisher, and Fedorenko]{blank_functional_2014}
Idan Blank, Nancy Kanwisher, and Evelina Fedorenko.
\newblock A functional dissociation between language and multiple-demand systems revealed in patterns of {BOLD} signal fluctuations.
\newblock \emph{Journal of Neurophysiology}, 112\penalty0 (5):\penalty0 1105--1118, September 2014.
\newblock ISSN 0022-3077, 1522-1598.
\newblock \doi{10.1152/jn.00884.2013}.
\newblock URL \url{https://www.physiology.org/doi/10.1152/jn.00884.2013}.

\bibitem[Bosselut et~al.(2019)Bosselut, Rashkin, Sap, Malaviya, Celikyilmaz, and Choi]{Bosselut2019COMETCT}
Antoine Bosselut, Hannah Rashkin, Maarten Sap, Chaitanya Malaviya, Asli Celikyilmaz, and Yejin Choi.
\newblock Comet: Commonsense transformers for automatic knowledge graph construction.
\newblock In \emph{Annual Meeting of the Association for Computational Linguistics}, 2019.
\newblock URL \url{https://api.semanticscholar.org/CorpusID:189762527}.

\bibitem[Caucheteux \& King(2022)Caucheteux and King]{caucheteux_brains_2022}
Charlotte Caucheteux and Jean-Rémi King.
\newblock Brains and algorithms partially converge in natural language processing.
\newblock \emph{Communications Biology}, 5\penalty0 (1):\penalty0 134, February 2022.
\newblock ISSN 2399-3642.
\newblock \doi{10.1038/s42003-022-03036-1}.
\newblock URL \url{https://www.nature.com/articles/s42003-022-03036-1}.

\bibitem[Chia et~al.(2023)Chia, Hong, Bing, and Poria]{chia_instructeval_2023}
Yew~Ken Chia, Pengfei Hong, Lidong Bing, and Soujanya Poria.
\newblock {INSTRUCTEVAL}: {Towards} {Holistic} {Evaluation} of {Instruction}-{Tuned} {Large} {Language} {Models}, June 2023.
\newblock URL \url{http://arxiv.org/abs/2306.04757}.
\newblock arXiv:2306.04757 [cs].

\bibitem[Chiang et~al.(2023)Chiang, Li, Lin, Sheng, Wu, Zhang, Zheng, Zhuang, Zhuang, Gonzalez, Stoica, and Xing]{vicuna2023}
Wei-Lin Chiang, Zhuohan Li, Zi~Lin, Ying Sheng, Zhanghao Wu, Hao Zhang, Lianmin Zheng, Siyuan Zhuang, Yonghao Zhuang, Joseph~E. Gonzalez, Ion Stoica, and Eric~P. Xing.
\newblock Vicuna: An open-source chatbot impressing gpt-4 with 90\%* chatgpt quality, March 2023.
\newblock URL \url{https://lmsys.org/blog/2023-03-30-vicuna/}.

\bibitem[Chung et~al.(2022)Chung, Hou, Longpre, Zoph, Tay, Fedus, Li, Wang, Dehghani, Brahma, Webson, Gu, Dai, Suzgun, Chen, Chowdhery, Castro-Ros, Pellat, Robinson, Valter, Narang, Mishra, Yu, Zhao, Huang, Dai, Yu, Petrov, Chi, Dean, Devlin, Roberts, Zhou, Le, and Wei]{chung_scaling_2022}
Hyung~Won Chung, Le~Hou, Shayne Longpre, Barret Zoph, Yi~Tay, William Fedus, Yunxuan Li, Xuezhi Wang, Mostafa Dehghani, Siddhartha Brahma, Albert Webson, Shixiang~Shane Gu, Zhuyun Dai, Mirac Suzgun, Xinyun Chen, Aakanksha Chowdhery, Alex Castro-Ros, Marie Pellat, Kevin Robinson, Dasha Valter, Sharan Narang, Gaurav Mishra, Adams Yu, Vincent Zhao, Yanping Huang, Andrew Dai, Hongkun Yu, Slav Petrov, Ed~H. Chi, Jeff Dean, Jacob Devlin, Adam Roberts, Denny Zhou, Quoc~V. Le, and Jason Wei.
\newblock Scaling {Instruction}-{Finetuned} {Language} {Models}, December 2022.
\newblock URL \url{http://arxiv.org/abs/2210.11416}.
\newblock arXiv:2210.11416 [cs].

\bibitem[Dapello et~al.(2020)Dapello, Marques, Schrimpf, Geiger, Cox, and DiCarlo]{dapello_simulating_2020}
Joel Dapello, Tiago Marques, Martin Schrimpf, Franziska Geiger, David~D. Cox, and James~J. DiCarlo.
\newblock Simulating a {{Primary Visual Cortex}} at the {{Front}} of {{CNNs Improves Robustness}} to {{Image Perturbations}}.
\newblock In \emph{Neural {{Information Processing Systems}} ({{NeurIPS}})}, June 2020.
\newblock \doi{10.1101/2020.06.16.154542}.

\bibitem[Dapello et~al.(2022)Dapello, Kar, Schrimpf, Geary, Ferguson, Cox, and DiCarlo]{dapello_aligning_2022}
Joel Dapello, Kohitij Kar, Martin Schrimpf, Robert Geary, Michael Ferguson, David~D. Cox, and James~J. DiCarlo.
\newblock Aligning {Model} and {Macaque} {Inferior} {Temporal} {Cortex} {Representations} {Improves} {Model}-to-{Human} {Behavioral} {Alignment} and {Adversarial} {Robustness}.
\newblock preprint, Neuroscience, July 2022.
\newblock URL \url{http://biorxiv.org/lookup/doi/10.1101/2022.07.01.498495}.

\bibitem[Dasgupta et~al.(2022)Dasgupta, Lampinen, Chan, Creswell, Kumaran, McClelland, and Hill]{dasgupta_language_2022}
Ishita Dasgupta, Andrew~K. Lampinen, Stephanie C.~Y. Chan, Antonia Creswell, Dharshan Kumaran, James~L. McClelland, and Felix Hill.
\newblock Language models show human-like content effects on reasoning, July 2022.
\newblock URL \url{http://arxiv.org/abs/2207.07051}.
\newblock arXiv:2207.07051 [cs].

\bibitem[Dong \& Toneva(2023)Dong and Toneva]{dong_interpreting_2023}
Dota~Tianai Dong and Mariya Toneva.
\newblock Interpreting multimodal video transformers using brain recordings.
\newblock In \emph{ICLR 2023 Workshop on Multimodal Representation Learning: Perks and Pitfalls}, 2023.
\newblock URL \url{https://openreview.net/forum?id=p-vL3rmYoqh}.

\bibitem[{Eghbal A. Hosseini} et~al.(2023){Eghbal A. Hosseini}, {Martin Schrimpf}, {Yian Zhang}, {Samuel Bowman}, {Noga Zaslavsky}, and {Evelina Fedorenko}]{eghbal_a_hosseini_artificial_2023}
{Eghbal A. Hosseini}, {Martin Schrimpf}, {Yian Zhang}, {Samuel Bowman}, {Noga Zaslavsky}, and {Evelina Fedorenko}.
\newblock Artificial neural network language models predict human brain responses to language even after a developmentally realistic amount of training.
\newblock \emph{bioRxiv}, pp.\  2022.10.04.510681, January 2023.
\newblock \doi{10.1101/2022.10.04.510681}.
\newblock URL \url{http://biorxiv.org/content/early/2023/09/19/2022.10.04.510681.abstract}.

\bibitem[Ehrlich \& Rayner(1981)Ehrlich and Rayner]{ehrlich_contextual_1981}
Susan~F. Ehrlich and Keith Rayner.
\newblock Contextual effects on word perception and eye movements during reading.
\newblock \emph{Journal of Verbal Learning and Verbal Behavior}, 20\penalty0 (6):\penalty0 641--655, December 1981.
\newblock ISSN 00225371.
\newblock \doi{10.1016/S0022-5371(81)90220-6}.
\newblock URL \url{https://linkinghub.elsevier.com/retrieve/pii/S0022537181902206}.

\bibitem[Futrell et~al.(2018)Futrell, Gibson, Tily, Blank, Vishnevetsky, Piantadosi, and Fedorenko]{futrell_natural_2018}
Richard Futrell, Edward Gibson, Harry~J. Tily, Idan Blank, Anastasia Vishnevetsky, Steven Piantadosi, and Evelina Fedorenko.
\newblock The natural stories corpus.
\newblock In \emph{Proceedings of the Eleventh International Conference on Language Resources and Evaluation ({LREC} 2018)}, Miyazaki, Japan, May 2018. European Language Resources Association (ELRA).
\newblock URL \url{https://aclanthology.org/L18-1012}.

\bibitem[Gao et~al.(2023)Gao, Huang, Li, and Chen]{gao-etal-2023-roles}
Changjiang Gao, Shujian Huang, Jixing Li, and Jiajun Chen.
\newblock Roles of scaling and instruction tuning in language perception: Model vs. human attention.
\newblock In Houda Bouamor, Juan Pino, and Kalika Bali (eds.), \emph{Findings of the Association for Computational Linguistics: EMNLP 2023}, pp.\  13042--13055, Singapore, December 2023. Association for Computational Linguistics.
\newblock \doi{10.18653/v1/2023.findings-emnlp.868}.
\newblock URL \url{https://aclanthology.org/2023.findings-emnlp.868}.

\bibitem[Goldstein et~al.(2022)Goldstein, Zada, Buchnik, Schain, Price, Aubrey, Nastase, Feder, Emanuel, Cohen, Jansen, Gazula, Choe, Rao, Kim, Casto, Fanda, Doyle, Friedman, Dugan, Melloni, Reichart, Devore, Flinker, Hasenfratz, Levy, Hassidim, Brenner, Matias, Norman, Devinsky, and Hasson]{goldstein_shared_2022}
Ariel Goldstein, Zaid Zada, Eliav Buchnik, Mariano Schain, Amy Price, Bobbi Aubrey, Samuel~A. Nastase, Amir Feder, Dotan Emanuel, Alon Cohen, Aren Jansen, Harshvardhan Gazula, Gina Choe, Aditi Rao, Catherine Kim, Colton Casto, Lora Fanda, Werner Doyle, Daniel Friedman, Patricia Dugan, Lucia Melloni, Roi Reichart, Sasha Devore, Adeen Flinker, Liat Hasenfratz, Omer Levy, Avinatan Hassidim, Michael Brenner, Yossi Matias, Kenneth~A. Norman, Orrin Devinsky, and Uri Hasson.
\newblock Shared computational principles for language processing in humans and deep language models.
\newblock \emph{Nature Neuroscience}, 25\penalty0 (3):\penalty0 369--380, March 2022.
\newblock ISSN 1097-6256, 1546-1726.
\newblock \doi{10.1038/s41593-022-01026-4}.
\newblock URL \url{https://www.nature.com/articles/s41593-022-01026-4}.

\bibitem[Hale(2001)]{hale_probabilistic_2001}
John Hale.
\newblock A probabilistic earley parser as a psycholinguistic model.
\newblock In \emph{Second meeting of the {North} {American} {Chapter} of the {Association} for {Computational} {Linguistics} on {Language} technologies 2001 - {NAACL} '01}, pp.\  1--8, Pittsburgh, Pennsylvania, 2001. Association for Computational Linguistics.
\newblock \doi{10.3115/1073336.1073357}.
\newblock URL \url{http://portal.acm.org/citation.cfm?doid=1073336.1073357}.

\bibitem[Hendrycks et~al.(2021)Hendrycks, Burns, Basart, Zou, Mazeika, Song, and Steinhardt]{hendrycks_measuring_2021}
Dan Hendrycks, Collin Burns, Steven Basart, Andy Zou, Mantas Mazeika, Dawn Song, and Jacob Steinhardt.
\newblock Measuring {Massive} {Multitask} {Language} {Understanding}, January 2021.
\newblock URL \url{http://arxiv.org/abs/2009.03300}.
\newblock arXiv:2009.03300 [cs].

\bibitem[Ivanova et~al.(2020)Ivanova, Srikant, Sueoka, Kean, Dhamala, O'Reilly, Bers, and Fedorenko]{ivanova_comprehension_2020}
Anna~A Ivanova, Shashank Srikant, Yotaro Sueoka, Hope~H Kean, Riva Dhamala, Una-May O'Reilly, Marina~U Bers, and Evelina Fedorenko.
\newblock Comprehension of computer code relies primarily on domain-general executive brain regions.
\newblock \emph{eLife}, 9:\penalty0 e58906, December 2020.
\newblock ISSN 2050-084X.
\newblock \doi{10.7554/eLife.58906}.
\newblock URL \url{https://elifesciences.org/articles/58906}.

\bibitem[Jain \& Huth(2018)Jain and Huth]{jain_incorporating_2018}
Shailee Jain and Alexander~G Huth.
\newblock Incorporating {Context} into {Language} {Encoding} {Models} for {fMRI}.
\newblock preprint, Neuroscience, May 2018.
\newblock URL \url{http://biorxiv.org/lookup/doi/10.1101/327601}.

\bibitem[Kauf et~al.(2024)Kauf, Chersoni, Lenci, Fedorenko, and Ivanova]{kauf2024comparing}
Carina Kauf, Emmanuele Chersoni, Alessandro Lenci, Evelina Fedorenko, and Anna~A. Ivanova.
\newblock Comparing plausibility estimates in base and instruction-tuned large language models, 2024.

\bibitem[Koo et~al.(2023)Koo, Lee, Raheja, Park, Kim, and Kang]{koo2023benchmarking}
Ryan Koo, Minhwa Lee, Vipul Raheja, Jong~Inn Park, Zae~Myung Kim, and Dongyeop Kang.
\newblock Benchmarking cognitive biases in large language models as evaluators, 2023.

\bibitem[Merlin \& Toneva(2022)Merlin and Toneva]{merlin_language_2022}
Gabriele Merlin and Mariya Toneva.
\newblock Language models and brain alignment: beyond word-level semantics and prediction, December 2022.
\newblock URL \url{http://arxiv.org/abs/2212.00596}.
\newblock arXiv:2212.00596 [cs, q-bio].

\bibitem[Oh \& Schuler(2023)Oh and Schuler]{oh_why_2023}
Byung-Doh Oh and William Schuler.
\newblock Why {Does} {Surprisal} {From} {Larger} {Transformer}-{Based} {Language} {Models} {Provide} a {Poorer} {Fit} to {Human} {Reading} {Times}?
\newblock \emph{Transactions of the Association for Computational Linguistics}, 11:\penalty0 336--350, March 2023.
\newblock ISSN 2307-387X.
\newblock \doi{10.1162/tacl_a_00548}.
\newblock URL \url{https://doi.org/10.1162/tacl_a_00548}.

\bibitem[Oh et~al.(2022)Oh, Clark, and Schuler]{oh_comparison_2022}
Byung-Doh Oh, Christian Clark, and William Schuler.
\newblock Comparison of {Structural} {Parsers} and {Neural} {Language} {Models} as {Surprisal} {Estimators}.
\newblock \emph{Frontiers in Artificial Intelligence}, 5:\penalty0 777963, March 2022.
\newblock ISSN 2624-8212.
\newblock \doi{10.3389/frai.2022.777963}.
\newblock URL \url{https://www.frontiersin.org/articles/10.3389/frai.2022.777963/full}.

\bibitem[Oota et~al.(2022)Oota, Arora, Agarwal, Marreddy, Gupta, and Surampudi]{oota_neural_2022}
Subba~Reddy Oota, Jashn Arora, Veeral Agarwal, Mounika Marreddy, Manish Gupta, and Bapi Surampudi.
\newblock Neural {Language} {Taskonomy}: {Which} {NLP} {Tasks} are the most {Predictive} of {fMRI} {Brain} {Activity}?
\newblock In \emph{Proceedings of the 2022 {Conference} of the {North} {American} {Chapter} of the {Association} for {Computational} {Linguistics}: {Human} {Language} {Technologies}}, pp.\  3220--3237, Seattle, United States, 2022. Association for Computational Linguistics.
\newblock \doi{10.18653/v1/2022.naacl-main.235}.
\newblock URL \url{https://aclanthology.org/2022.naacl-main.235}.

\bibitem[Oota et~al.(2023)Oota, Gupta, Bapi, Jobard, Alexandre, and Hinaut]{oota_deep_2023}
Subba~Reddy Oota, Manish Gupta, Raju~S. Bapi, Gael Jobard, Frederic Alexandre, and Xavier Hinaut.
\newblock Deep {Neural} {Networks} and {Brain} {Alignment}: {Brain} {Encoding} and {Decoding} ({Survey}), July 2023.
\newblock URL \url{http://arxiv.org/abs/2307.10246}.
\newblock arXiv:2307.10246 [cs, q-bio].

\bibitem[Ouyang et~al.(2022)Ouyang, Wu, Jiang, Almeida, Wainwright, Mishkin, Zhang, Agarwal, Slama, Ray, Schulman, Hilton, Kelton, Miller, Simens, Askell, Welinder, Christiano, Leike, and Lowe]{Ouyang2022TrainingLM}
Long Ouyang, Jeff Wu, Xu~Jiang, Diogo Almeida, Carroll~L. Wainwright, Pamela Mishkin, Chong Zhang, Sandhini Agarwal, Katarina Slama, Alex Ray, John Schulman, Jacob Hilton, Fraser Kelton, Luke~E. Miller, Maddie Simens, Amanda Askell, Peter Welinder, Paul~Francis Christiano, Jan Leike, and Ryan~J. Lowe.
\newblock Training language models to follow instructions with human feedback.
\newblock \emph{ArXiv}, abs/2203.02155, 2022.
\newblock URL \url{https://api.semanticscholar.org/CorpusID:246426909}.

\bibitem[Pereira et~al.(2018)Pereira, Lou, Pritchett, Ritter, Gershman, Kanwisher, Botvinick, and Fedorenko]{pereira_toward_2018}
Francisco Pereira, Bin Lou, Brianna Pritchett, Samuel Ritter, Samuel~J. Gershman, Nancy Kanwisher, Matthew Botvinick, and Evelina Fedorenko.
\newblock Toward a universal decoder of linguistic meaning from brain activation.
\newblock \emph{Nature Communications}, 9\penalty0 (1):\penalty0 963, March 2018.
\newblock ISSN 2041-1723.
\newblock \doi{10.1038/s41467-018-03068-4}.
\newblock URL \url{https://www.nature.com/articles/s41467-018-03068-4}.

\bibitem[Raffel et~al.(2020)Raffel, Shazeer, Roberts, Lee, Narang, Matena, Zhou, Li, and Liu]{raffel_exploring_2020}
Colin Raffel, Noam Shazeer, Adam Roberts, Katherine Lee, Sharan Narang, Michael Matena, Yanqi Zhou, Wei Li, and Peter~J. Liu.
\newblock Exploring the {Limits} of {Transfer} {Learning} with a {Unified} {Text}-to-{Text} {Transformer}, July 2020.
\newblock URL \url{http://arxiv.org/abs/1910.10683}.
\newblock arXiv:1910.10683 [cs, stat].

\bibitem[Rowling et~al.(1998)Rowling, GrandPre, GrandPr{\'e}, Taylor, Books, and Inc]{rowling1998harry}
J.K. Rowling, M.~GrandPre, M.~GrandPr{\'e}, T.~Taylor, Arthur A.~Levine Books, and Scholastic Inc.
\newblock \emph{Harry Potter and the Sorcerer's Stone}.
\newblock Harry Potter. A.A. Levine Books, 1998.
\newblock ISBN 9780590353403.
\newblock URL \url{https://books.google.de/books?id=zXgTdQagLGkC}.

\bibitem[Safarani et~al.(2021)Safarani, Nix, Willeke, Cadena, Restivo, Denfield, Tolias, and Sinz]{safarani_towards_2021}
Shahd Safarani, Arne Nix, Konstantin Willeke, Santiago Cadena, Kelli Restivo, George Denfield, Andreas Tolias, and Fabian Sinz.
\newblock Towards robust vision by multi-task learning on monkey visual cortex.
\newblock \emph{Advances in Neural Information Processing Systems}, 34:\penalty0 739--751, 2021.

\bibitem[Safdari et~al.(2023)Safdari, Serapio-García, Crepy, Fitz, Romero, Sun, Abdulhai, Faust, and Matarić]{safdari_personality_2023}
Mustafa Safdari, Greg Serapio-García, Clément Crepy, Stephen Fitz, Peter Romero, Luning Sun, Marwa Abdulhai, Aleksandra Faust, and Maja Matarić.
\newblock Personality {Traits} in {Large} {Language} {Models}, June 2023.
\newblock URL \url{http://arxiv.org/abs/2307.00184}.
\newblock arXiv:2307.00184 [cs].

\bibitem[Sap et~al.(2020)Sap, Shwartz, Bosselut, Choi, and Roth]{sap_commonsense_2020}
Maarten Sap, Vered Shwartz, Antoine Bosselut, Yejin Choi, and Dan Roth.
\newblock Commonsense {Reasoning} for {Natural} {Language} {Processing}.
\newblock In \emph{Proceedings of the 58th {Annual} {Meeting} of the {Association} for {Computational} {Linguistics}: {Tutorial} {Abstracts}}, pp.\  27--33, Online, 2020. Association for Computational Linguistics.
\newblock \doi{10.18653/v1/2020.acl-tutorials.7}.
\newblock URL \url{https://www.aclweb.org/anthology/2020.acl-tutorials.7}.

\bibitem[Schrimpf et~al.(2018)Schrimpf, Kubilius, Hong, Majaj, Rajalingham, Issa, Kar, Bashivan, Prescott-Roy, Geiger, Schmidt, Yamins, and DiCarlo]{schrimpf_brain-score_2018}
Martin Schrimpf, Jonas Kubilius, Ha~Hong, Najib~J. Majaj, Rishi Rajalingham, Elias~B. Issa, Kohitij Kar, Pouya Bashivan, Jonathan Prescott-Roy, Franziska Geiger, Kailyn Schmidt, Daniel L.~K. Yamins, and James~J. DiCarlo.
\newblock Brain-{Score}: {Which} {Artificial} {Neural} {Network} for {Object} {Recognition} is most {Brain}-{Like}?
\newblock preprint, Neuroscience, September 2018.
\newblock URL \url{http://biorxiv.org/lookup/doi/10.1101/407007}.

\bibitem[Schrimpf et~al.(2020)Schrimpf, Kubilius, Lee, Ratan~Murty, Ajemian, and DiCarlo]{schrimpf_integrative_2020}
Martin Schrimpf, Jonas Kubilius, Michael~J Lee, N.~Apurva Ratan~Murty, Robert Ajemian, and James~J. DiCarlo.
\newblock Integrative {{Benchmarking}} to {{Advance Neurally Mechanistic Models}} of {{Human Intelligence}}.
\newblock \emph{Neuron}, 2020.
\newblock ISSN 0896-6273.
\newblock \doi{10.1016/j.neuron.2020.07.040}.

\bibitem[Schrimpf et~al.(2021)Schrimpf, Blank, Tuckute, Kauf, Hosseini, Kanwisher, Tenenbaum, and Fedorenko]{schrimpf_neural_2021}
Martin Schrimpf, Idan~Asher Blank, Greta Tuckute, Carina Kauf, Eghbal~A. Hosseini, Nancy Kanwisher, Joshua~B. Tenenbaum, and Evelina Fedorenko.
\newblock The neural architecture of language: {Integrative} modeling converges on predictive processing.
\newblock \emph{Proceedings of the National Academy of Sciences}, 118\penalty0 (45):\penalty0 e2105646118, November 2021.
\newblock ISSN 0027-8424, 1091-6490.
\newblock \doi{10.1073/pnas.2105646118}.
\newblock URL \url{https://pnas.org/doi/full/10.1073/pnas.2105646118}.

\bibitem[Schwartz et~al.(2019)Schwartz, Toneva, and Wehbe]{schwartz_inducing_2019}
Dan Schwartz, Mariya Toneva, and Leila Wehbe.
\newblock Inducing brain-relevant bias in natural language processing models, October 2019.
\newblock URL \url{http://arxiv.org/abs/1911.03268}.
\newblock arXiv:1911.03268 [cs, q-bio].

\bibitem[Smith \& Levy(2013)Smith and Levy]{smith_effect_2013}
Nathaniel~J. Smith and Roger Levy.
\newblock The effect of word predictability on reading time is logarithmic.
\newblock \emph{Cognition}, 128\penalty0 (3):\penalty0 302--319, 2013.
\newblock ISSN 0010-0277.
\newblock \doi{https://doi.org/10.1016/j.cognition.2013.02.013}.
\newblock URL \url{https://www.sciencedirect.com/science/article/pii/S0010027713000413}.

\bibitem[Steuer et~al.(2023)Steuer, Mosbach, and Klakow]{steuer-etal-2023-large}
Julius Steuer, Marius Mosbach, and Dietrich Klakow.
\newblock Large {GPT}-like models are bad babies: A closer look at the relationship between linguistic competence and psycholinguistic measures.
\newblock In Alex Warstadt, Aaron Mueller, Leshem Choshen, Ethan Wilcox, Chengxu Zhuang, Juan Ciro, Rafael Mosquera, Bhargavi Paranjabe, Adina Williams, Tal Linzen, and Ryan Cotterell (eds.), \emph{Proceedings of the BabyLM Challenge at the 27th Conference on Computational Natural Language Learning}, pp.\  142--157, Singapore, December 2023. Association for Computational Linguistics.
\newblock \doi{10.18653/v1/2023.conll-babylm.12}.
\newblock URL \url{https://aclanthology.org/2023.conll-babylm.12}.

\bibitem[Suzgun et~al.(2022)Suzgun, Scales, Schärli, Gehrmann, Tay, Chung, Chowdhery, Le, Chi, Zhou, and Wei]{suzgun_challenging_2022}
Mirac Suzgun, Nathan Scales, Nathanael Schärli, Sebastian Gehrmann, Yi~Tay, Hyung~Won Chung, Aakanksha Chowdhery, Quoc~V. Le, Ed~H. Chi, Denny Zhou, and Jason Wei.
\newblock Challenging {BIG}-{Bench} {Tasks} and {Whether} {Chain}-of-{Thought} {Can} {Solve} {Them}, October 2022.
\newblock URL \url{http://arxiv.org/abs/2210.09261}.
\newblock arXiv:2210.09261 [cs].

\bibitem[Taori et~al.(2023)Taori, Gulrajani, Zhang, Dubois, Li, Guestrin, Liang, and Hashimoto]{alpaca}
Rohan Taori, Ishaan Gulrajani, Tianyi Zhang, Yann Dubois, Xuechen Li, Carlos Guestrin, Percy Liang, and Tatsunori~B. Hashimoto.
\newblock Stanford alpaca: An instruction-following llama model.
\newblock \url{https://github.com/tatsu-lab/stanford_alpaca}, 2023.

\bibitem[Team et~al.(2024)Team, Mesnard, Hardin, Dadashi, Bhupatiraju, Pathak, Sifre, Rivière, Kale, Love, Tafti, Hussenot, Sessa, Chowdhery, Roberts, Barua, Botev, Castro-Ros, Slone, Héliou, Tacchetti, Bulanova, Paterson, Tsai, Shahriari, Lan, Choquette-Choo, Crepy, Cer, Ippolito, Reid, Buchatskaya, Ni, Noland, Yan, Tucker, Muraru, Rozhdestvenskiy, Michalewski, Tenney, Grishchenko, Austin, Keeling, Labanowski, Lespiau, Stanway, Brennan, Chen, Ferret, Chiu, Mao-Jones, Lee, Yu, Millican, Sjoesund, Lee, Dixon, Reid, Mikuła, Wirth, Sharman, Chinaev, Thain, Bachem, Chang, Wahltinez, Bailey, Michel, Yotov, Chaabouni, Comanescu, Jana, Anil, McIlroy, Liu, Mullins, Smith, Borgeaud, Girgin, Douglas, Pandya, Shakeri, De, Klimenko, Hennigan, Feinberg, Stokowiec, hui Chen, Ahmed, Gong, Warkentin, Peran, Giang, Farabet, Vinyals, Dean, Kavukcuoglu, Hassabis, Ghahramani, Eck, Barral, Pereira, Collins, Joulin, Fiedel, Senter, Andreev, and Kenealy]{gemmateam2024gemmaopenmodelsbased}
Gemma Team, Thomas Mesnard, Cassidy Hardin, Robert Dadashi, Surya Bhupatiraju, Shreya Pathak, Laurent Sifre, Morgane Rivière, Mihir~Sanjay Kale, Juliette Love, Pouya Tafti, Léonard Hussenot, Pier~Giuseppe Sessa, Aakanksha Chowdhery, Adam Roberts, Aditya Barua, Alex Botev, Alex Castro-Ros, Ambrose Slone, Amélie Héliou, Andrea Tacchetti, Anna Bulanova, Antonia Paterson, Beth Tsai, Bobak Shahriari, Charline~Le Lan, Christopher~A. Choquette-Choo, Clément Crepy, Daniel Cer, Daphne Ippolito, David Reid, Elena Buchatskaya, Eric Ni, Eric Noland, Geng Yan, George Tucker, George-Christian Muraru, Grigory Rozhdestvenskiy, Henryk Michalewski, Ian Tenney, Ivan Grishchenko, Jacob Austin, James Keeling, Jane Labanowski, Jean-Baptiste Lespiau, Jeff Stanway, Jenny Brennan, Jeremy Chen, Johan Ferret, Justin Chiu, Justin Mao-Jones, Katherine Lee, Kathy Yu, Katie Millican, Lars~Lowe Sjoesund, Lisa Lee, Lucas Dixon, Machel Reid, Maciej Mikuła, Mateo Wirth, Michael Sharman, Nikolai Chinaev, Nithum Thain, Olivier Bachem,
  Oscar Chang, Oscar Wahltinez, Paige Bailey, Paul Michel, Petko Yotov, Rahma Chaabouni, Ramona Comanescu, Reena Jana, Rohan Anil, Ross McIlroy, Ruibo Liu, Ryan Mullins, Samuel~L Smith, Sebastian Borgeaud, Sertan Girgin, Sholto Douglas, Shree Pandya, Siamak Shakeri, Soham De, Ted Klimenko, Tom Hennigan, Vlad Feinberg, Wojciech Stokowiec, Yu~hui Chen, Zafarali Ahmed, Zhitao Gong, Tris Warkentin, Ludovic Peran, Minh Giang, Clément Farabet, Oriol Vinyals, Jeff Dean, Koray Kavukcuoglu, Demis Hassabis, Zoubin Ghahramani, Douglas Eck, Joelle Barral, Fernando Pereira, Eli Collins, Armand Joulin, Noah Fiedel, Evan Senter, Alek Andreev, and Kathleen Kenealy.
\newblock Gemma: Open models based on gemini research and technology, 2024.
\newblock URL \url{https://arxiv.org/abs/2403.08295}.

\bibitem[Toneva \& Wehbe(2019)Toneva and Wehbe]{toneva_interpreting_2019}
Mariya Toneva and Leila Wehbe.
\newblock Interpreting and improving natural-language processing (in machines) with natural language-processing (in the brain), November 2019.
\newblock URL \url{http://arxiv.org/abs/1905.11833}.
\newblock arXiv:1905.11833 [cs, q-bio].

\bibitem[Touvron et~al.(2023{\natexlab{a}})Touvron, Lavril, Izacard, Martinet, Lachaux, Lacroix, Rozière, Goyal, Hambro, Azhar, Rodriguez, Joulin, Grave, and Lample]{touvron_llama_2023}
Hugo Touvron, Thibaut Lavril, Gautier Izacard, Xavier Martinet, Marie-Anne Lachaux, Timothée Lacroix, Baptiste Rozière, Naman Goyal, Eric Hambro, Faisal Azhar, Aurelien Rodriguez, Armand Joulin, Edouard Grave, and Guillaume Lample.
\newblock {LLaMA}: {Open} and {Efficient} {Foundation} {Language} {Models}, February 2023{\natexlab{a}}.
\newblock URL \url{http://arxiv.org/abs/2302.13971}.
\newblock arXiv:2302.13971 [cs].

\bibitem[Touvron et~al.(2023{\natexlab{b}})Touvron, Martin, Stone, Albert, Almahairi, Babaei, Bashlykov, Batra, Bhargava, Bhosale, Bikel, Blecher, Ferrer, Chen, Cucurull, Esiobu, Fernandes, Fu, Fu, Fuller, Gao, Goswami, Goyal, Hartshorn, Hosseini, Hou, Inan, Kardas, Kerkez, Khabsa, Kloumann, Korenev, Koura, Lachaux, Lavril, Lee, Liskovich, Lu, Mao, Martinet, Mihaylov, Mishra, Molybog, Nie, Poulton, Reizenstein, Rungta, Saladi, Schelten, Silva, Smith, Subramanian, Tan, Tang, Taylor, Williams, Kuan, Xu, Yan, Zarov, Zhang, Fan, Kambadur, Narang, Rodriguez, Stojnic, Edunov, and Scialom]{touvron2023llama2openfoundation}
Hugo Touvron, Louis Martin, Kevin Stone, Peter Albert, Amjad Almahairi, Yasmine Babaei, Nikolay Bashlykov, Soumya Batra, Prajjwal Bhargava, Shruti Bhosale, Dan Bikel, Lukas Blecher, Cristian~Canton Ferrer, Moya Chen, Guillem Cucurull, David Esiobu, Jude Fernandes, Jeremy Fu, Wenyin Fu, Brian Fuller, Cynthia Gao, Vedanuj Goswami, Naman Goyal, Anthony Hartshorn, Saghar Hosseini, Rui Hou, Hakan Inan, Marcin Kardas, Viktor Kerkez, Madian Khabsa, Isabel Kloumann, Artem Korenev, Punit~Singh Koura, Marie-Anne Lachaux, Thibaut Lavril, Jenya Lee, Diana Liskovich, Yinghai Lu, Yuning Mao, Xavier Martinet, Todor Mihaylov, Pushkar Mishra, Igor Molybog, Yixin Nie, Andrew Poulton, Jeremy Reizenstein, Rashi Rungta, Kalyan Saladi, Alan Schelten, Ruan Silva, Eric~Michael Smith, Ranjan Subramanian, Xiaoqing~Ellen Tan, Binh Tang, Ross Taylor, Adina Williams, Jian~Xiang Kuan, Puxin Xu, Zheng Yan, Iliyan Zarov, Yuchen Zhang, Angela Fan, Melanie Kambadur, Sharan Narang, Aurelien Rodriguez, Robert Stojnic, Sergey Edunov, and Thomas
  Scialom.
\newblock Llama 2: Open foundation and fine-tuned chat models, 2023{\natexlab{b}}.
\newblock URL \url{https://arxiv.org/abs/2307.09288}.

\bibitem[van Duijn et~al.(2023)van Duijn, van Dijk, Kouwenhoven, de~Valk, Spruit, and van~der Putten]{van-duijn-etal-2023-theory}
Max van Duijn, Bram van Dijk, Tom Kouwenhoven, Werner de~Valk, Marco Spruit, and Peter van~der Putten.
\newblock Theory of mind in large language models: Examining performance of 11 state-of-the-art models vs. children aged 7-10 on advanced tests.
\newblock In Jing Jiang, David Reitter, and Shumin Deng (eds.), \emph{Proceedings of the 27th Conference on Computational Natural Language Learning (CoNLL)}, pp.\  389--402, Singapore, December 2023. Association for Computational Linguistics.
\newblock \doi{10.18653/v1/2023.conll-1.25}.
\newblock URL \url{https://aclanthology.org/2023.conll-1.25}.

\bibitem[Wang et~al.(2022{\natexlab{a}})Wang, Kordi, Mishra, Liu, Smith, Khashabi, and Hajishirzi]{selfinstruct}
Yizhong Wang, Yeganeh Kordi, Swaroop Mishra, Alisa Liu, Noah~A. Smith, Daniel Khashabi, and Hannaneh Hajishirzi.
\newblock Self-instruct: Aligning language model with self generated instructions, 2022{\natexlab{a}}.

\bibitem[Wang et~al.(2022{\natexlab{b}})Wang, Mishra, Alipoormolabashi, Kordi, Mirzaei, Arunkumar, Ashok, Dhanasekaran, Naik, Stap, Pathak, Karamanolakis, Lai, Purohit, Mondal, Anderson, Kuznia, Doshi, Patel, Pal, Moradshahi, Parmar, Purohit, Varshney, Kaza, Verma, Puri, Karia, Sampat, Doshi, Mishra, Reddy, Patro, Dixit, Shen, Baral, Choi, Smith, Hajishirzi, and Khashabi]{wang_super-naturalinstructions_2022}
Yizhong Wang, Swaroop Mishra, Pegah Alipoormolabashi, Yeganeh Kordi, Amirreza Mirzaei, Anjana Arunkumar, Arjun Ashok, Arut~Selvan Dhanasekaran, Atharva Naik, David Stap, Eshaan Pathak, Giannis Karamanolakis, Haizhi~Gary Lai, Ishan Purohit, Ishani Mondal, Jacob Anderson, Kirby Kuznia, Krima Doshi, Maitreya Patel, Kuntal~Kumar Pal, Mehrad Moradshahi, Mihir Parmar, Mirali Purohit, Neeraj Varshney, Phani~Rohitha Kaza, Pulkit Verma, Ravsehaj~Singh Puri, Rushang Karia, Shailaja~Keyur Sampat, Savan Doshi, Siddhartha Mishra, Sujan Reddy, Sumanta Patro, Tanay Dixit, Xudong Shen, Chitta Baral, Yejin Choi, Noah~A. Smith, Hannaneh Hajishirzi, and Daniel Khashabi.
\newblock Super-{NaturalInstructions}: {Generalization} via {Declarative} {Instructions} on 1600+ {NLP} {Tasks}, October 2022{\natexlab{b}}.
\newblock URL \url{http://arxiv.org/abs/2204.07705}.
\newblock arXiv:2204.07705 [cs].

\bibitem[Wang et~al.(2022{\natexlab{c}})Wang, Mishra, Alipoormolabashi, Kordi, Mirzaei, Naik, Ashok, Dhanasekaran, Arunkumar, Stap, Pathak, Karamanolakis, Lai, Purohit, Mondal, Anderson, Kuznia, Doshi, Pal, Patel, Moradshahi, Parmar, Purohit, Varshney, Kaza, Verma, Puri, Karia, Doshi, Sampat, Mishra, Reddy~A, Patro, Dixit, and Shen]{wang-etal-2022-super}
Yizhong Wang, Swaroop Mishra, Pegah Alipoormolabashi, Yeganeh Kordi, Amirreza Mirzaei, Atharva Naik, Arjun Ashok, Arut~Selvan Dhanasekaran, Anjana Arunkumar, David Stap, Eshaan Pathak, Giannis Karamanolakis, Haizhi Lai, Ishan Purohit, Ishani Mondal, Jacob Anderson, Kirby Kuznia, Krima Doshi, Kuntal~Kumar Pal, Maitreya Patel, Mehrad Moradshahi, Mihir Parmar, Mirali Purohit, Neeraj Varshney, Phani~Rohitha Kaza, Pulkit Verma, Ravsehaj~Singh Puri, Rushang Karia, Savan Doshi, Shailaja~Keyur Sampat, Siddhartha Mishra, Sujan Reddy~A, Sumanta Patro, Tanay Dixit, and Xudong Shen.
\newblock Super-{N}atural{I}nstructions: Generalization via declarative instructions on 1600+ {NLP} tasks.
\newblock In \emph{Proceedings of the 2022 Conference on Empirical Methods in Natural Language Processing}, pp.\  5085--5109, Abu Dhabi, United Arab Emirates, December 2022{\natexlab{c}}. Association for Computational Linguistics.
\newblock \doi{10.18653/v1/2022.emnlp-main.340}.
\newblock URL \url{https://aclanthology.org/2022.emnlp-main.340}.

\bibitem[Wehbe et~al.(2014)Wehbe, Murphy, Talukdar, Fyshe, Ramdas, and Mitchell]{wehbe_simultaneously_2014}
Leila Wehbe, Brian Murphy, Partha Talukdar, Alona Fyshe, Aaditya Ramdas, and Tom Mitchell.
\newblock Simultaneously {Uncovering} the {Patterns} of {Brain} {Regions} {Involved} in {Different} {Story} {Reading} {Subprocesses}.
\newblock \emph{PLoS ONE}, 9\penalty0 (11):\penalty0 e112575, November 2014.
\newblock ISSN 1932-6203.
\newblock \doi{10.1371/journal.pone.0112575}.
\newblock URL \url{https://dx.plos.org/10.1371/journal.pone.0112575}.

\bibitem[Wilcox et~al.(2020)Wilcox, Gauthier, Hu, Qian, and Levy]{wilcox_predictive_2020}
Ethan~Gotlieb Wilcox, Jon Gauthier, Jennifer Hu, Peng Qian, and Roger Levy.
\newblock On the {Predictive} {Power} of {Neural} {Language} {Models} for {Human} {Real}-{Time} {Comprehension} {Behavior}, June 2020.
\newblock URL \url{http://arxiv.org/abs/2006.01912}.
\newblock arXiv:2006.01912 [cs].

\bibitem[Zhang et~al.(2023)Zhang, Dong, Li, Zhang, Sun, Wang, Li, Hu, Zhang, Wu, and Wang]{zhang_instruction_2023}
Shengyu Zhang, Linfeng Dong, Xiaoya Li, Sen Zhang, Xiaofei Sun, Shuhe Wang, Jiwei Li, Runyi Hu, Tianwei Zhang, Fei Wu, and Guoyin Wang.
\newblock Instruction {Tuning} for {Large} {Language} {Models}: {A} {Survey}, August 2023.
\newblock URL \url{http://arxiv.org/abs/2308.10792}.
\newblock arXiv:2308.10792 [cs].

\end{thebibliography}
\bibliographystyle{colm2024_conference}

\appendix
\clearpage
\section{Language Models: Parameter count and Number of Layers}
\label{appendix_LM_counts}

\begin{table}[!ht]
    \centering
    \begin{tabular}{l r r}
        \toprule
        Model & Parameter Count & Number of Layers \\ 
        \midrule
        t5-small & 77 M & 16 \\ 
        flan-t5-small & 77 M & 16 \\ 
        t5-base & 250 M & 24 \\ 
        flan-t5-base & 250 M & 24 \\ 
        flan-alpaca-base & 250 M & 24 \\ 
        t5-large & 800 M & 48 \\ 
        flan-t5-large & 800 M & 48 \\ 
        flan-alpaca-large & 800 M & 48 \\ 
        t5-xl & 3 B & 48 \\ 
        flan-t5-xl & 3 B & 48 \\ 
        flan-alpaca-xl & 3 B & 48 \\ 
        flan-gpt4all-xl & 3 B & 48 \\ 
        flan-sharegpt-xl & 3 B & 48 \\ 
        flan-alpaca-gpt4-xl & 3 B & 48 \\ 
        t5-xxl & 11 B & 48 \\ 
        flan-t5-xxl & 11 B & 48 \\ 
        flan-alpaca-xxl & 11 B & 48 \\ 
        \midrule
        llama-7b & 7 B & 32 \\ 
        alpaca-7b & 7 B & 32 \\ 
        vicuna-7b & 7 B & 32 \\ 
        llama-13b & 13 B & 40 \\ 
        vicuna-13b & 13 B & 40 \\ 
        stable-vicuna-13b & 13 B & 40 \\ 
        llama-33b & 33 B & 60 \\ 
        vicuna-33b & 33 B & 60 \\ 
        \midrule
        gpt2-small & 124 M & 12 \\
        gpt2-small-alpaca & 124 M & 12 \\
        gpt2-medium & 355 M & 24 \\
        gpt2-medium-alpaca & 355 M & 24 \\
        gpt2-large & 774 M & 36 \\
        gpt2-large-alpaca & 774 M & 36 \\
        gpt2-xl & 1.5 B & 48 \\
        gpt2-xl-alpaca & 1.5 B & 48 \\
        \bottomrule
    \end{tabular}
    \caption{
        \textbf{Parameter count and number of layers for all vanilla and instruction-tuned LLMs.}
        For the parameter count, ``M" refers to million and ``B" refers to billion.
        The number of layers for T5 models is a sum of the number of encoder and decoder layers.
    }
\end{table}

\clearpage
\section{Language Models: Links to models weights}
\label{appendix_LM_urls}

\begin{table}[!ht]
    \centering
    \begin{tabular}{l l}
        \toprule
        Model & Link to model weights \\ 
        \midrule
        t5-small & \url{www.huggingface.co/google/t5-v1\_1-small} \\ 
        flan-t5-small & \url{www.huggingface.co/google/flan-t5-small} \\ 
        t5-base & \url{www.huggingface.co/google/t5-v1\_1-base} \\ 
        flan-t5-base & \url{www.huggingface.co/google/flan-t5-base} \\ 
        flan-alpaca-base & \url{www.huggingface.co/declare-lab/flan-alpaca-base} \\ 
        t5-large & \url{www.huggingface.co/google/t5-v1\_1-large} \\ 
        flan-t5-large & \url{www.huggingface.co/google/flan-t5-large} \\ 
        flan-alpaca-large & \url{www.huggingface.co/declare-lab/flan-alpaca-large} \\ 
        t5-xl & \url{www.huggingface.co/google/t5-v1\_1-xl} \\ 
        flan-t5-xl & \url{www.huggingface.co/google/flan-t5-xl} \\ 
        flan-alpaca-xl & \url{www.huggingface.co/declare-lab/flan-alpaca-xl} \\ 
        flan-gpt4all-xl & \url{www.huggingface.co/declare-lab/flan-gpt4all-xl} \\ 
        flan-sharegpt-xl & \url{www.huggingface.co/declare-lab/flan-sharegpt-xl} \\ 
        flan-alpaca-gpt4-xl & \url{www.huggingface.co/declare-lab/flan-alpaca-gpt4-xl} \\ 
        t5-xxl & \url{www.huggingface.co/google/t5-v1\_1-xxl} \\ 
        flan-t5-xxl & \url{www.huggingface.co/google/flan-t5-xxl} \\ 
        flan-alpaca-xxl & \url{www.huggingface.co/declare-lab/flan-alpaca-xxl} \\ 
        \midrule
        llama-7b & \url{www.github.com/facebookresearch/llama} \\ 
        alpaca-7b & \url{www.github.com/tatsu-lab/stanford\_alpaca} \\ 
        vicuna-7b & \url{www.huggingface.co/lmsys/vicuna-7b-v1.3} \\ 
        llama-13b & \url{www.github.com/facebookresearch/llama} \\ 
        vicuna-13b & \url{www.huggingface.co/lmsys/vicuna-13b-v1.3} \\ 
        stable-vicuna-13b & \url{www.huggingface.co/CarperAI/stable-vicuna-13b-delta} \\ 
        llama-33b & \url{www.github.com/facebookresearch/llama} \\ 
        vicuna-33b & \url{www.huggingface.co/lmsys/vicuna-33b-v1.3} \\ 
        \midrule
        gpt2-small & \url{https://huggingface.co/openai-community/gpt2} \\
        gpt2-small-alpaca & \url{https://huggingface.co/vicgalle/gpt2-alpaca} \\
        gpt2-medium & \url{https://huggingface.co/openai-community/gpt2-medium} \\
        gpt2-medium-alpaca & \url{https://huggingface.co/linkanjarad/GPT2-Medium-Alpaca-355m} \\
        gpt2-large & \url{https://huggingface.co/openai-community/gpt2-large} \\
        gpt2-large-alpaca & \url{https://huggingface.co/reasonwang/gpt2-large-alpaca} \\
        gpt2-xl & \url{https://huggingface.co/openai-community/gpt2-xl} \\
        gpt2-xl-alpaca & \url{https://huggingface.co/Rachneet/gpt2-xl-alpaca} \\
        \bottomrule
    \end{tabular}
    \caption{
        \textbf{Link to model weights for all vanilla and instruction-tuned LLMs.}
        We provide these links for reproducibility purposes.
    }
\end{table}

\clearpage
\section{Method for computing Brain alignment}
\label{appendix_methods_brain_alignment}

\begin{figure}[h]
    \includegraphics[width=1.0\linewidth]{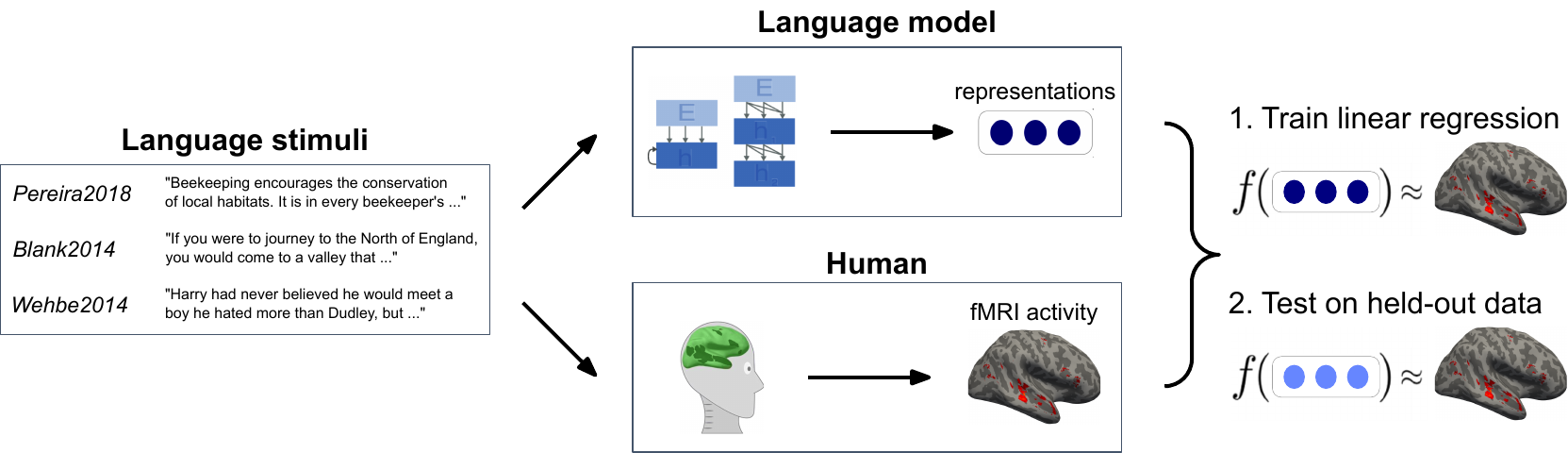}
    \caption{
        \textbf{Method for computing Brain alignment}, the similarity of an LLM’s internal representations to human brain activity. 
    }
    \label{fig:appendix_brain_alignment_method}
\end{figure}

We follow a general approach for aligning LLM representations to fMRI brain activity used in prior works \citep{jain_incorporating_2018, toneva_interpreting_2019, schrimpf_neural_2021}. The approaches used for the three neural datasets are similar. Hence, we explain the main details only for \dataset{Wehbe2014}, followed by key differences for the other two datasets.

\subsection{\dataset{Wehbe2014}}
\dataset{Wehbe2014} \citep{wehbe_simultaneously_2014}: The data includes fMRI recordings of 8 human participants reading chapter 9 of the book \textit{Harry Potter and the Sorceror's Stone} \citep{rowling1998harry}. Participants read the chapter at a fixed interval of one word every 0.5 seconds. The chapter contains 5,176 words.

\textbf{LLM representations.} We extract LLM representations of the text at each layer $\in \RR^{w\times d}$, where $w$ is the number of words in the text and $d$ is the embedding size of the LLM.

\textbf{fMRI activity.}
Data analyses were performed on fMRI BOLD signals extracted from the language network for all three neural datasets.
fMRI activity was recorded at regular intervals of 2 seconds, with a total of 1,351 recorded time-points (TRs, times of repetition).
We construct a stimulus-response matrix for fMRI activity $\in \RR^{TR \times v_i}$, where $TR$ is the number of TRs and $v_i$ is the number of voxels in the brain of participant $i$.

\textbf{Alignment of LLM representations to fMRI activity.} We use PCA to reduce the dimensionality of the LLM representations to result in a matrix $\in \RR^{w\times 10}$. As the number of words in the text is greater than the number of TRs, we down-sample the LLM word-level representations to the TR rate by averaging the LLM representations to the corresponding TRs, producing a matrix $\in \RR^{TR\times 10}$. As the response recorded by fMRI peaks about 6 seconds after stimulus onset, we follow prior methods by including preceding time-points for each time-point. The final LLM representations are constructed by concatenating, for each TR, the LLM representations corresponding to the previous 4 TRs, producing a matrix $\in \RR^{TR\times 40}$. This accounts for the lag in the hemodynamic response that fMRI records. Finally, we learn a linear function, regularized by the ridge penalty, that uses the modified LLM representations $\in \RR^{TR\times 40}$ to predict the fMRI activity of participants $\in \RR^{TR \times v_i}$. We train the function using 4-fold cross-validation, where each fold corresponds to a separate run of fMRI collection. We also remove 10 TRs from the LLM representations and fMRI activity from the ends of each test run to avoid train-test overlap. To compute the brain alignment, we evaluate the Pearson correlation (r) between the predicted fMRI activity $\in \RR^{TR \times v_i}$ and recorded fMRI activity $\in \RR^{TR \times v_i}$.

\subsection{\dataset{Blank2014}}
\dataset{Blank2014} \citep{blank_functional_2014}: The data consists of fMRI recordings of 5 human participants listening to 8 naturalistic stories from the Natural Stories Corpus \citep{futrell_natural_2018}.

fMRI activity was recorded at regular intervals of 2 seconds, with a total of 1,317 TRs. We average the BOLD signals across voxels within each language-responsive region of interest (ROI) of each participant to increase the signal-to-noise ratio, following \citet{schrimpf_neural_2021}. As there are 60 ROIs in total across the 5 participants, this produces a stimulus-response matrix $\in \RR^{1,317 \times 60}$.

\subsection{\dataset{Pereira2018}}
\dataset{Pereira2018} (experiments 2 and 3 from \citealp{pereira_toward_2018}): In experiment 2, 9 participants read 384 sentences taken from 96 text passages. In experiment 3, 6 participants read 243 sentences from 72 text passages. Each sentence was displayed for 4 seconds on a screen.

We average the fMRI responses for each sentence, resulting in one data point per sentence per language-responsive voxel of each participant. For experiment 2, there are 384 sentences and 12,195 language-responsive voxels across the participants. For experiment 3, there are 243 sentences and 8,121 language-responsive voxels across the participants. We concatenate the fMRI responses across sentences and participants for each experiment. This produces a stimulus-response matrix $\in \RR^{384 \times 12,195}$ for experiment 2 and a stimulus-response matrix $\in \RR^{243 \times 8121}$ for experiment 3.

\section{Method for computing Behavioral alignment}
\label{appendix_methods_behavioral_alignment}

\begin{figure}[h]
    \includegraphics[width=1.0\linewidth]{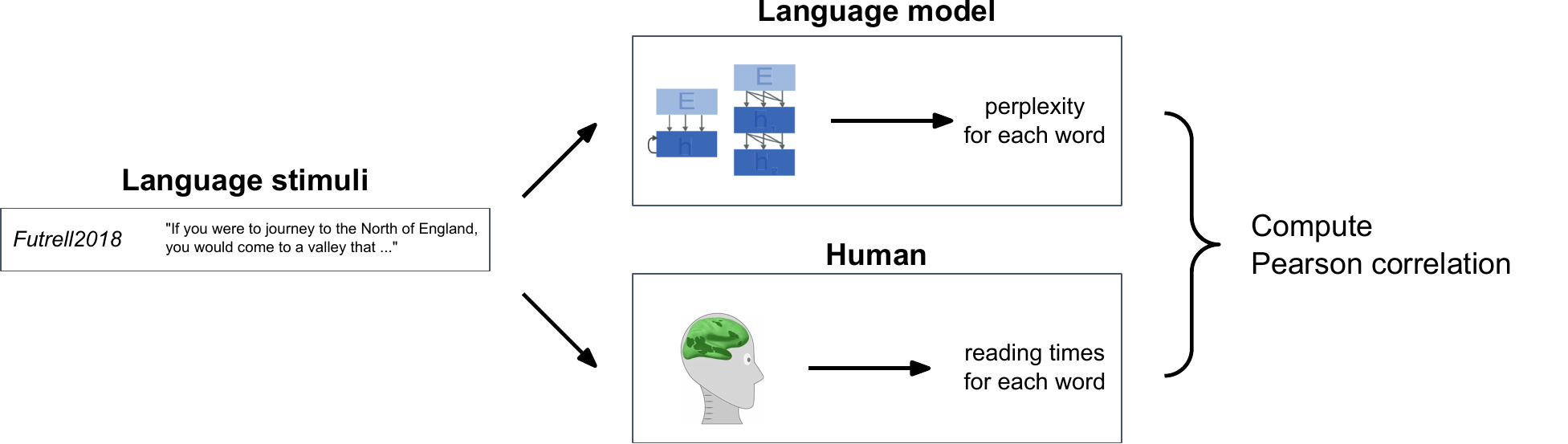}
    \caption{
        \textbf{Method for computing Behavioral alignment.} The same language stimuli are presented to LLMs and human participants, using the \dataset{Futrell2018} benchmark in Brain-Score, which contains naturalistic stories. We compute the behavioral alignment as the Pearson correlation between LLM perplexity for each word and human reading times for each word. 
    }
    \label{fig:appendix_behavioral_alignment_method}
\end{figure}

\dataset{Futrell2018} \citep{futrell_natural_2018}: The data consists of reading times (RTs) of 179 participants on a total of 10,256 words across 485 sentences, taken from 10 stories from the Natural Stories Corpus \citep{futrell_natural_2018}. The stories were presented online to Amazon Mechanical Turk users visually in a dashed moving window display. Participants press a button to reveal each consecutive word. When they press the button, the most recent word is converted back to dashes, and the next word is revealed. The time between button presses offers an estimate of comprehension difficulty \citep{ehrlich_contextual_1981, hale_probabilistic_2001, smith_effect_2013}. We construct a stimulus-response matrix of RTs $\in \RR^{10,256 \times 179}$.

We provide the same stimuli to an LLM and record its perplexity for each word, producing an LLM perplexity matrix $\in \RR^{10,256}$. We compute behavioral alignment as the Pearson correlation between per-word LLM perplexity and human RTs. Perplexity for LLMs and reading times for humans offer insights into comprehension difficulty \citep{ehrlich_contextual_1981, hale_probabilistic_2001, smith_effect_2013}, allowing us to examine whether LLMs and humans share similarities in terms of which words and sentences they find challenging or surprising.

\clearpage
\section{MMLU and BBH benchmarks}
\label{appendix_mmlu_bbh}

MMLU is designed to measure knowledge from many domains \cite{hendrycks_measuring_2021}. It contains 57 tasks, categorized by the subject domain of world knowledge tested: STEM, Humanities, Social Sciences, and Others. The STEM category includes questions on computer science, physics, mathematics, etc. The Humanities category includes questions on philosophy, law, history, etc. The Social Sciences category includes questions on politics, sociology, economics, geography, etc. The Others category includes questions on business topics such as finance, accounting, as well as general knowledge of global facts.

BBH is designed to evaluate various problem-solving and reasoning abilities of LLMs \citep{suzgun_challenging_2022}.
BBH contains 23 tasks, categorized by the type of problem-solving ability tested: (1) Algorithmic and Multi-Step Arithmetic Reasoning, (2) Natural Language Understanding, (3) Use of World Knowledge, and (4) Multilingual Knowledge and Reasoning. 
The world knowledge category of BBH contains tasks that test for factual and general knowledge. Tasks requiring factual knowledge include: “Sports Understanding” and “Movie Recommendation”. Tasks requiring general knowledge include: “Causal Judgement”, which tests knowledge about causal-reasoning suppositions, and “Ruin Names”, which requires knowledge about human perception and usage of humor in the English language.

For both benchmarks, we adopt the same category classification as used in their original papers. We measure the performance of LLMs on BBH and MMLU using the \texttt{instruct-eval} repository\footnote{\url{https://github.com/declare-lab/instruct-eval}} with default settings (3-shots, 5-shots respectively) and preset prompts.

\section{Code Repositories}
\label{appendix_code_repos}

We use the Brain-Score repository to evaluate brain alignment for the \dataset{Pereira2018} and \dataset{Blank2014} datasets, as well as behavioral alignment for the \dataset{Futrell2018} dataset. Link: \url{www.github.com/brain-score/language}. 

We use an open-source repository to evaluate brain alignment for the \dataset{Wehbe2014} dataset. Link: \url{www.github.com/awwkl/brain_language_summarization}, which builds on \url{www.github.com/mtoneva/brain_language_nlp}.

We use the Instruct-Eval repository to evaluate MMLU and BBH scores. Link: \url{www.github.com/declare-lab/instruct-eval}. 

We use the Stanford Alpaca repository for instruction-tuning. Link: \url{www.github.com/tatsu-lab/stanford_alpaca}).

\clearpage
\section{Results for Brain alignment}
\label{appendix_results_BA}

\begin{table}[!ht]
    \centering
    \begin{tabular}{l r r r r}
        \toprule
        \textbf{} & \dataset{Pereira2018} & \dataset{Blank2014} & \dataset{Wehbe2014} & Average \\
        \midrule
        t5-small & 0.166 & 0.168 & 0.071 & 0.135 \\ 
        flan-t5-small & 0.202 & 0.178 & 0.079 & 0.153 \\ 
        t5-base & 0.222 & 0.188 & 0.074 & 0.162 \\ 
        flan-t5-base & 0.234 & 0.178 & 0.076 & 0.163 \\ 
        flan-alpaca-base & 0.227 & 0.179 & 0.076 & 0.161 \\ 
        t5-large & 0.270 & 0.082 & 0.071 & 0.141 \\ 
        flan-t5-large & 0.311 & 0.104 & 0.080 & 0.165 \\ 
        flan-alpaca-large & 0.322 & 0.126 & 0.082 & 0.177 \\ 
        t5-xl & 0.285 & 0.192 & 0.072 & 0.183 \\ 
        flan-t5-xl & 0.314 & 0.215 & 0.072 & 0.200 \\ 
        flan-alpaca-xl & 0.312 & 0.209 & 0.075 & 0.199 \\ 
        flan-gpt4all-xl & 0.300 & 0.206 & 0.078 & 0.195 \\ 
        flan-sharegpt-xl & 0.323 & 0.211 & 0.070 & 0.201 \\ 
        flan-alpaca-gpt4-xl & 0.302 & 0.205 & 0.073 & 0.193 \\ 
        t5-xxl & 0.343 & 0.297 & 0.096 & 0.246 \\ 
        flan-t5-xxl & 0.350 & 0.268 & 0.103 & 0.240 \\ 
        flan-alpaca-xxl & 0.346 & 0.268 & 0.102 & 0.239 \\ 
        \midrule
        llama-7b & 0.405 & 0.154 & 0.118 & 0.226 \\ 
        alpaca-7b & 0.420 & 0.167 & 0.118 & 0.235 \\ 
        vicuna-7b & 0.399 & 0.152 & 0.119 & 0.223 \\ 
        llama-13b & 0.412 & 0.133 & 0.115 & 0.220 \\ 
        vicuna-13b & 0.423 & 0.148 & 0.116 & 0.229 \\ 
        stable-vicuna-13b & 0.415 & 0.144 & 0.115 & 0.225 \\ 
        llama-33b & 0.426 & 0.145 & 0.109 & 0.227 \\ 
        vicuna-33b & 0.436 & 0.156 & 0.105 & 0.232 \\ 
        \midrule
        gpt2-small & 0.305 & 0.123 & 0.088 & 0.172 \\
        gpt2-small-alpaca & 0.298 & 0.121 & 0.081 & 0.167 \\
        gpt2-medium & 0.329 & 0.081 & 0.089 & 0.166 \\
        gpt2-medium-alpaca & 0.325 & 0.080 & 0.090 & 0.165 \\
        gpt2-large & 0.342 & 0.077 & 0.101 & 0.173 \\
        gpt2-large-alpaca & 0.336 & 0.074 & 0.101 & 0.170 \\
        gpt2-xl & 0.358 & 0.140 & 0.102 & 0.200 \\
        gpt2-xl-alpaca & 0.343 & 0.139 & 0.093 & 0.192 \\
        \bottomrule
    \end{tabular}
    \caption{
        \textbf{Brain alignment results for all vanilla and instruction-tuned LLMs.} 
        We provide these results for reproducibility purposes.
    }
\end{table}

\begin{table}[!ht]
    \centering
    \begin{tabular}{l r r r r}
        \toprule
        \textbf{} & \dataset{Pereira2018} & \dataset{Blank2014} & \dataset{Wehbe2014} & Average \\
        \midrule
        Noise ceiling & 0.359  & 0.210  & 0.104 &  0.224 \\ 
        \bottomrule
    \end{tabular}
    \caption{
        \textbf{Noise ceiling estimates for all 3 neural datasets.} 
        \camera{
        fMRI measurements inherently include noise, i.e., fluctuations not due to neurons firing, from sources such as environmental factors, the scanner, and movements of human participants. Consequently, a ``noise ceiling" is often computed by recording multiple samples from the same subjects for the same stimuli, providing an upper-limit estimate of the possible correlation in fMRI activity. In our study, we followed prior work for calculating the noise ceiling. 
        }
        For \dataset{Pereira2018} and \dataset{Blank2014}, noise ceiling estimates are computed using the Brain-Score repository, with details provided in \citet{schrimpf_neural_2021}. For \dataset{Wehbe2014}, noise ceiling estimates are computed using a similar procedure.
    }
\end{table}


\clearpage
\section{Results for Next-word prediction, MMLU, BBH}
\label{appendix_results_mmlu_bbh}

\begin{table}[!ht]
    \centering
    \begin{tabular}{l r r r}
        \toprule
         & WikiText-2 NWP Loss & MMLU Score & BBH Score \\ 
        \midrule
        flan-t5-small & 0.851 & 0.294 & 0.287 \\ 
        flan-t5-base & 1.235 & 0.341 & 0.308 \\ 
        flan-alpaca-base & 1.074 & 0.304 & \textcolor{gray}{0.266} \\ 
        flan-t5-large & 0.625 & 0.419 & 0.370 \\ 
        flan-alpaca-large & 0.648 & 0.397 & \textcolor{gray}{0.276} \\ 
        flan-t5-xl & 0.650 & 0.493 & 0.402 \\ 
        flan-alpaca-xl & 0.604 & 0.466 & \textcolor{gray}{0.270} \\ 
        flan-gpt4all-xl & 0.625 & 0.337 & \textcolor{gray}{0.212} \\ 
        flan-sharegpt-xl & 0.664 & 0.446 & 0.363 \\ 
        flan-alpaca-gpt4-xl & 0.593 & 0.456 & 0.348 \\ 
        flan-t5-xxl & 0.638 & 0.545 & 0.443 \\ 
        flan-alpaca-xxl & 0.607 & 0.508 & \textcolor{gray}{0.229} \\ 
        \midrule
        alpaca-7b & 4.201 & 0.404 & 0.328 \\ 
        vicuna-7b & 4.387 & 0.472 & 0.331 \\ 
        vicuna-13b & 4.130 & 0.521 & 0.387 \\ 
        stable-vicuna-13b & 4.623 & 0.495 & 0.380 \\ 
        vicuna-33b & 3.940 & 0.590 & 0.426 \\ 
        \midrule
        gpt2-small-alpaca & 4.193 & \textcolor{gray}{0.270} & \textcolor{gray}{0.273} \\ 
        gpt2-medium-alpaca & 3.333 & \textcolor{gray}{0.260} & \textcolor{gray}{0.276} \\
        gpt2-large-alpaca & 3.614 & \textcolor{gray}{0.260} & \textcolor{gray}{0.278} \\
        gpt2-xl-alpaca & 3.283 & \textcolor{gray}{0.270} & \textcolor{gray}{0.284} \\ 
        \bottomrule
    \end{tabular}
    \caption{
        \textbf{WikiText-2 NWP loss, MMLU Overall Score, and BBH Overall Score for all instruction-tuned LLMs.} 
        Results for vanilla LLMs are not shown as they are not adapted for the question formats in the MMLU and BBH benchmarks. Results in \textcolor{gray}{gray} are close to random performance. We provide these results for reproducibility purposes.
    }
\end{table}

\textbf{Notes on comparing next-word prediction (NWP) loss across model families.} \: The T5 and LLaMA models belong to separate model families. We wish to caution that comparing next-word prediction loss across different model families may not be meaningful. This is due to several reasons related to architectural differences, training methodologies, and objectives.
(1) Architecture: T5 models have an encoder-decoder architecture while LLaMA models have a decoder-only architecture.
(2) Training Objectives: The T5 models were trained on supervised and unsupervised tasks, while the LLaMA models were trained only on unsupervised text (Section \ref{method_language_models}).
(3) Loss computation: The loss functions for both model families are computed differently, making it inappropriate to directly compare their loss values.
(4) Evaluation Metrics: Next-word prediction loss is just one metric, and it may not capture the overall language understanding capabilities of a model. Hence, we additionally evaluate these LLMs' alignment to human brain activity, as well as their performance on problem-solving abilities (BBH) and tasks requiring world knowledge (MMLU).
In summary, while NWP loss is a valuable metric for evaluating language models within the same family or architecture, comparing across different model families may not be meaningful.

\clearpage
\section{Results for Correlations of Brain Alignment with LLM properties}
\label{appendix_results_corrs_BA}

\begin{figure}[h]
    \centering
    \includegraphics[width=1.0\linewidth]{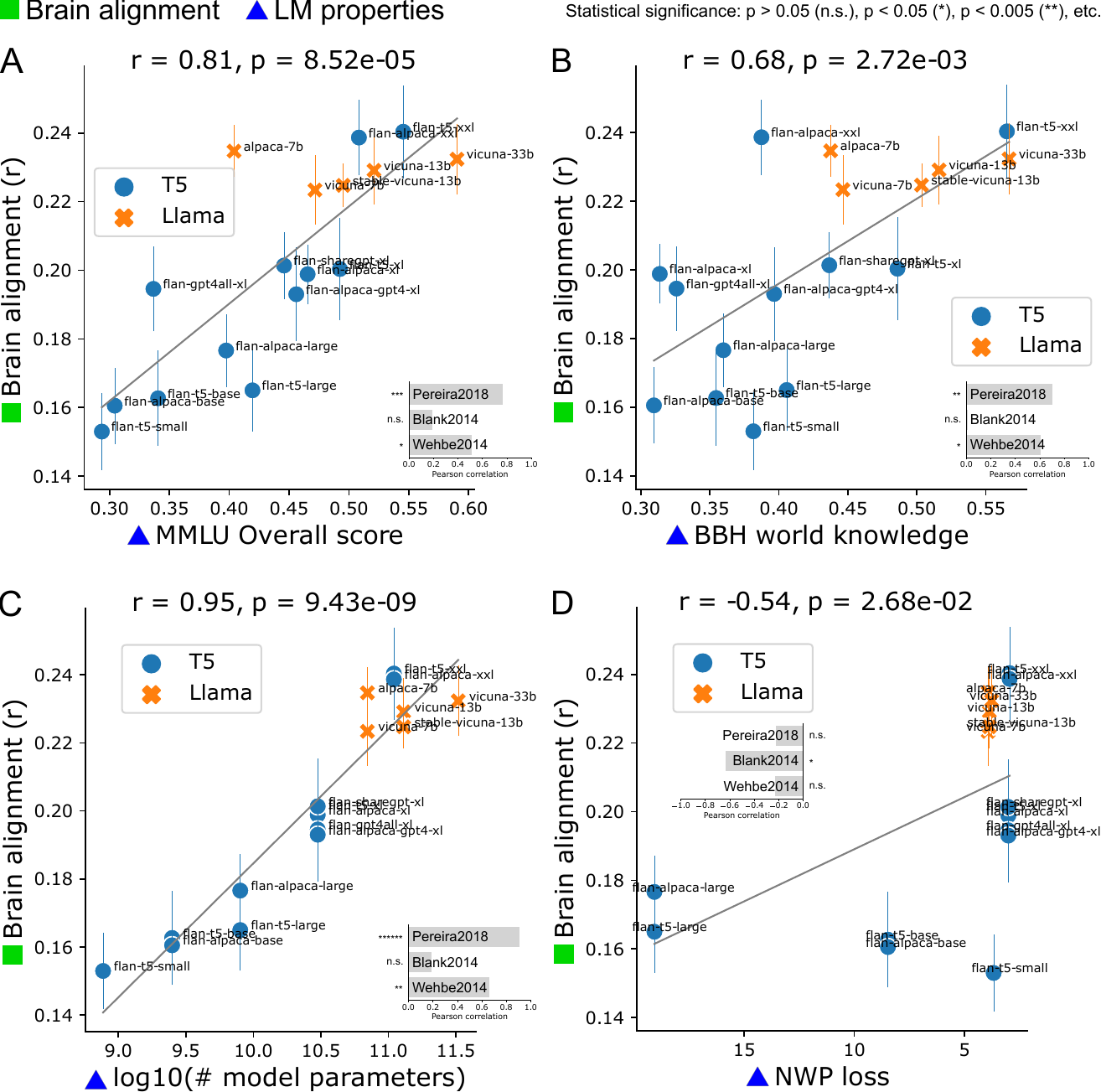}
    \caption{
        \textbf{Correlation between brain alignment and various LLM properties:} (A) MMLU benchmark overall score, (B) BBH benchmark score for world knowledge tasks, (C) number of parameters of the model, and (D) Next-word prediction (NWP) loss.}
    \label{fig:appendix_corrs_BA}
\end{figure}

\clearpage
\begin{figure}[h]
    \centering
    \includegraphics[width=1.0\linewidth]{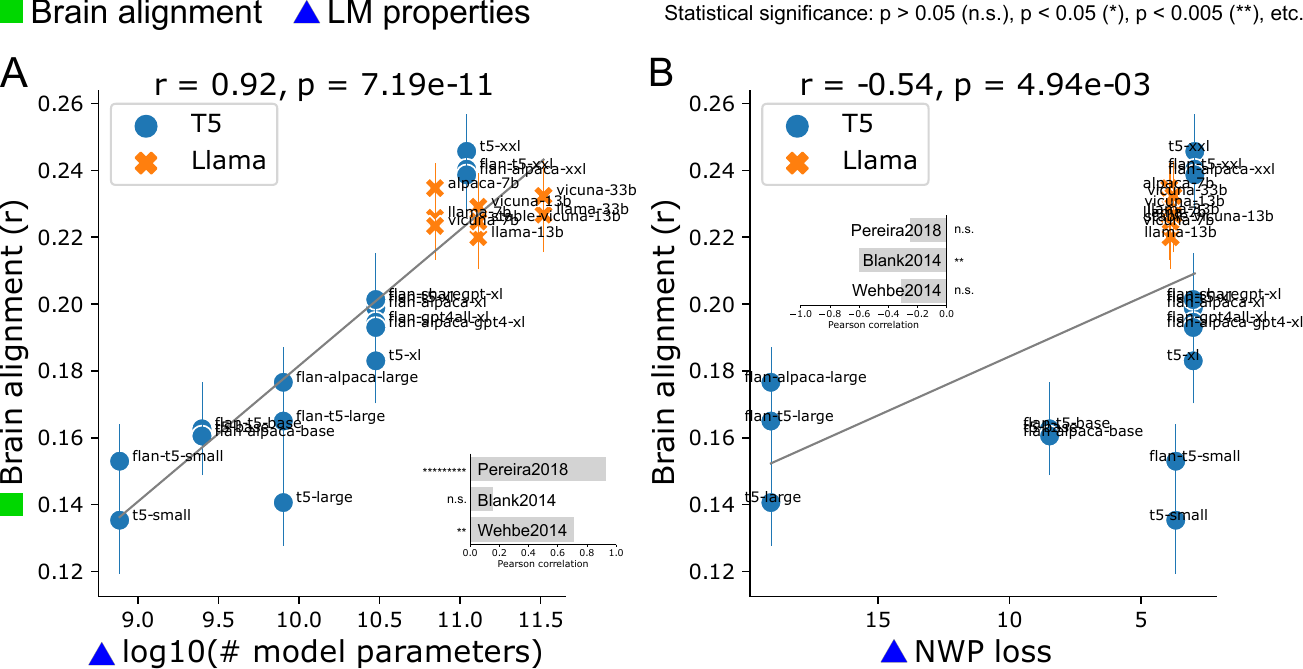}
    \caption{
        \textbf{Correlation between brain alignment and various LLM properties for all LLMs (including vanilla models):} (A) number of parameters of the model, and (B) Next-word prediction (NWP) loss.}
    \label{fig:appendix_corrs_BA_all}
\end{figure}

\camera{
\section{Additional Brain Alignment results (Linear, RSA, CKA) for Gemma and LLaMA-2 models}
\label{appendix_ba_results_rsa_cka}

\begin{table}[h!]
    \centering
        \begin{minipage}{.47\linewidth}
            \begin{tabular}{l c c c}
            \toprule
            \multicolumn{4}{c}{Pretrained models} \\
            \midrule
            \textbf{Model} & \textbf{Linear} & \textbf{CKA} & \textbf{RSA} \\
            \midrule
            Gemma-2B & 0.76 & 1.94 & 0.63 \\
            Gemma-7B & 0.90 & 2.25 & 0.62 \\
            LLaMA-2-7B & 0.97 & 1.81 & 0.65 \\
            LLaMA-2-13B & 1.08 & 2.51 & 0.68 \\
            \bottomrule
            \end{tabular}
        \end{minipage}%
    \hfill
        \begin{minipage}{.47\linewidth}
            \begin{tabular}{l c c c}
            \toprule
            \multicolumn{4}{c}{Instruction-tuned models} \\
            \midrule
            \textbf{Model} & \textbf{Linear} & \textbf{CKA} & \textbf{RSA} \\
            \midrule
            Gemma-2B & 0.72 & 2.44 & 0.70 \\
            Gemma-7B & 0.94 & 2.55 & 0.72 \\
            LLaMA-2-7B & 1.02 & 2.00 & 0.63 \\
            LLaMA-2-13B & 1.08 & 2.77 & 0.65 \\
            \bottomrule
            \end{tabular}
        \end{minipage}%
    \hfill
    \caption{\textbf{Additional Brain Alignment results (Linear, RSA, CKA) for Gemma and LLaMA-2 models.} ``Linear" refers to the linear predictivity metric used in the main paper, normalized by the noise ceiling computed in Appendix \ref{appendix_results_BA}.}
    \label{table:appendix_ba_results_rsa_cka}
\end{table}

In Section \ref{section_brain_alignment}, we study the representational similarity between LLMs and human brain activity (brain alignment) using the linear predictivity similarity metric (``Linear"). This metric involves training a linear regression model to predict fMRI brain activity based on LLM representations. Here, we additionally validate our results using the RSA and CKA similarity metrics (Table \ref{table:appendix_ba_results_rsa_cka}), confirming that the brain alignment results are consistent across different similarity metrics.

In Section \ref{subsection_result_IT_BA}, we observed that instruction-tuning generally improves brain alignment for models from the T5 and LLaMA-1 families. We further validate this trend with recently released models from the Gemma and LLaMA-2 families (Table \ref{table:appendix_ba_results_rsa_cka}). These results hold across various similarity metrics, including Linear predictivity, RSA, and CKA.

}

\clearpage
\section{Results for Instruction-tuning LLaMA-7B on Alpaca dataset}
\label{appendix_results_IT_llama_7b}

\begin{figure}[h]
    \includegraphics[width=1.0\linewidth]{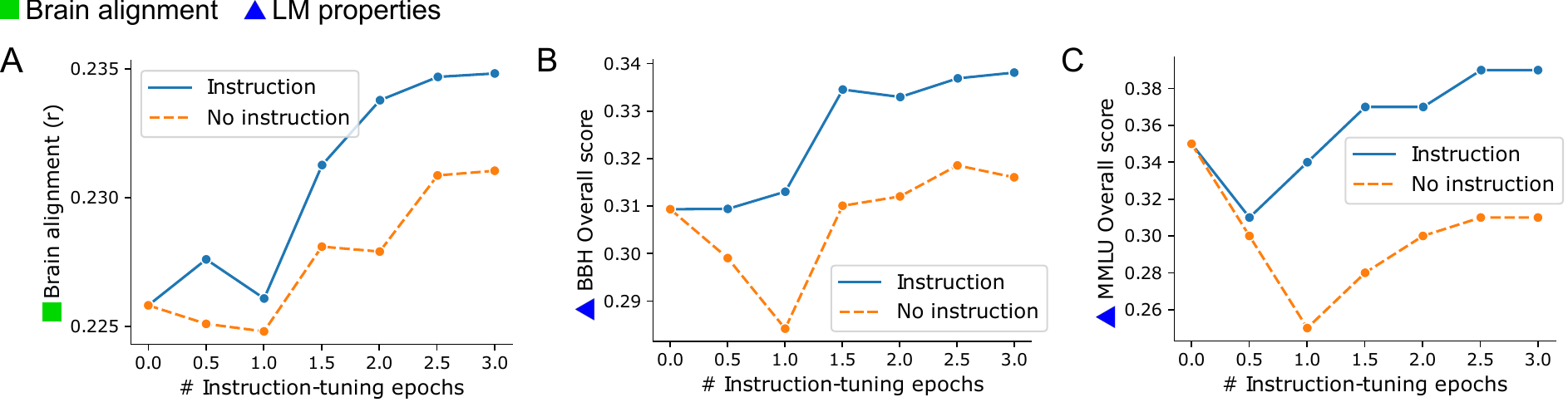}
    \caption{
        \textbf{Improvements in brain alignment from instruction-tuning are due to both additional training data, as well as training to understand and follow instructions.}
    }
    \label{fig:appendix_instruct_alpaca}
\end{figure}

\paragraph{Instruction model.} We instruction-tune LLaMA-7B on the Stanford Alpaca dataset \citep{alpaca} using the default training process, following the code in \url{www.github.com/tatsu-lab/stanford_alpaca}. In particular, the model is instruction-tuned using $52$K instruction-following examples generated through methods inspired by Self-Instruct \citep{selfinstruct}. This model is labeled ``Instruction" in Figure \ref{fig:appendix_instruct_alpaca}. 

\paragraph{No instruction model (Ablation).} We also train an ablation model with the same process and training data as the default instruction-tuning, but remove the instruction portion from each training sample. This ablation model is labeled ``No instruction" in Figure \ref{fig:appendix_instruct_alpaca}. 
This ablation experiment disentangles: (1) training data (present in both ``Instruction" and ``No instruction"), from (2) training LLMs to understand and follow instructions (present only in ``Instruction").

We use all provided training samples from the Alpaca dataset, thus ensuring that the models are trained on the same amount of data. We observe that brain alignment of the ``No Instruction" ablation model increases during fine-tuning but stays lower than its ``Instruction" counterpart. This shows that brain alignment improvements are due to both (1) training data (present in both models) and (2) the process of training LLMs to understand and follow instructions (present only in the ``Instruction" model).

\clearpage
\section{Results for Behavioral alignment}
\label{appendix_results_BehavA}

\begin{table}[!ht]
    \centering
    \begin{tabular}{lr}
        \toprule
                            & \dataset{Futrell2018} \\
        \midrule
        t5-small            & 0.229       \\
        flan-t5-small       & 0.054       \\
        t5-base             & 0.333       \\
        flan-t5-base        & 0.152       \\
        flan-alpaca-base    & 0.290       \\
        t5-large            & 0.303       \\
        flan-t5-large       & 0.145       \\
        flan-alpaca-large   & 0.291       \\
        t5-xl               & 0.225       \\
        flan-t5-xl          & 0.113       \\
        flan-alpaca-xl      & 0.181       \\
        flan-gpt4all-xl     & 0.251       \\
        flan-sharegpt-xl    & 0.285       \\
        flan-alpaca-gpt4-xl & 0.250       \\
        t5-xxl              & 0.260       \\
        flan-t5-xxl         & 0.274       \\
        flan-alpaca-xxl     & 0.267       \\
        \midrule
        llama-7b            & 0.204       \\
        alpaca-7b           & 0.206       \\
        vicuna-7b           & 0.205       \\
        llama-13b           & 0.184       \\
        vicuna-13b          & 0.184       \\
        stable-vicuna-13b   & 0.196       \\
        llama-33b           & 0.158       \\
        vicuna-33b          & 0.164       \\
        \midrule
        gpt2-small & 0.367 \\
        gpt2-small-alpaca & 0.335 \\
        gpt2-medium & 0.350 \\
        gpt2-medium-alpaca & 0.345 \\
        gpt2-large & 0.331 \\
        gpt2-large-alpaca & 0.338 \\
        gpt2-xl & 0.318 \\
        gpt2-xl-alpaca & 0.336 \\
        \bottomrule
    \end{tabular}
    \caption{
        \textbf{Behavioral alignment results for all vanilla and instruction-tuned LLMs.} 
        We provide these results for reproducibility purposes.
    }
\end{table}

\begin{table}[!ht]
    \centering
    \begin{tabular}{l r}
        \toprule
        \textbf{} & \dataset{Futrell2018} \\
        \midrule
        Noise ceiling & 0.858 \\ 
        \bottomrule
    \end{tabular}
    \caption{
        \textbf{Noise ceiling estimates for the \dataset{Futrell2018} reading-times dataset.} Noise ceiling estimates are computed using the Brain-Score repository, with details provided in \citet{schrimpf_neural_2021}.
    }
\end{table}

\clearpage
\section{Results for Correlations of Behavioral Alignment with LLM properties}
\label{appendix_results_corrs_BehavA}

\begin{figure}[h]
    \centering
    \includegraphics[width=1.0\linewidth]{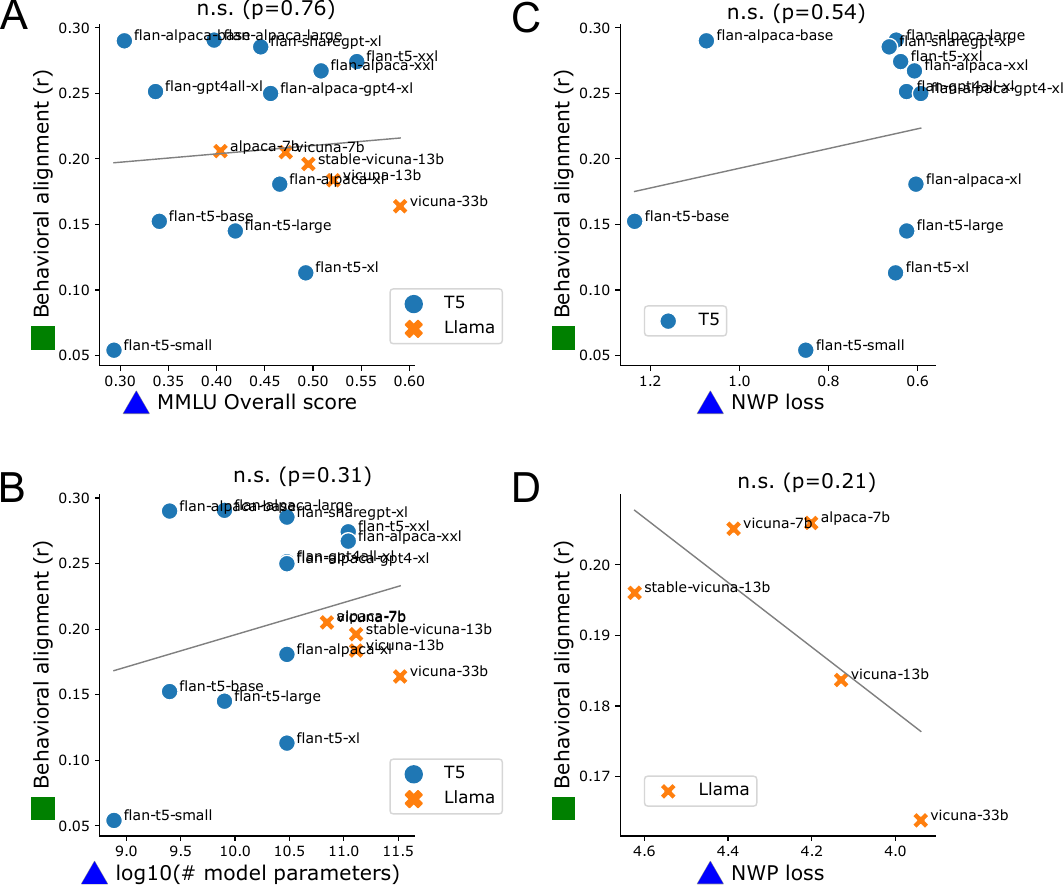}
    \caption{
        \textbf{Correlation between behavioral alignment and various LLM properties:} (A) MMLU benchmark overall score, (B) number of parameters of the model, (C) Next-word prediction (NWP) loss for T5 models, and (D) NWP loss for LLaMA models.}
    \label{fig:appendix_corrs_BehavA}
\end{figure}

\end{document}